\documentclass{article}

\usepackage{microtype}
\usepackage{graphicx}
\usepackage{subfigure}
\usepackage{booktabs} 
\usepackage[margin=1in]{geometry}
\usepackage{hyperref}
\usepackage{nicefrac}       
\usepackage{microtype}      
\usepackage{bm}
\usepackage{bbm}
\usepackage{graphicx}
\usepackage{amssymb,amsmath}
\usepackage[dvipsnames]{xcolor}
\usepackage{comment}

\usepackage{algorithm} 
\usepackage{algpseudocode}

\newcommand{\grad}{\nabla}
\renewcommand{\div}{\nabla\cdot}

\def\thetab{\boldsymbol{\theta}}
\def\Thetab{\boldsymbol{\Theta}}

\def\xb{\boldsymbol{x}}

\def\yb{\boldsymbol{y}}

\def\eps{\varepsilon}

\def\RR{\mathbb{R}} \def\NN{\mathbb{N}} 
\def\EE{\mathbb{E}}

\newtheorem{theorem}{Theorem}[section]
\newtheorem{proposition}[theorem]{Proposition}
\newtheorem{lemma}[theorem]{Lemma}
\newtheorem{assumption}[theorem]{Assumption}

\def\<{\langle} \def\>{\rangle}

\DeclareMathOperator{\supp}{supp}
\DeclareMathOperator{\sign}{sign}
\DeclareMathOperator{\tr}{tr}
\DeclareMathOperator{\argmin}{argmin}

\usepackage{authblk}
\title{Global convergence of neuron birth-death dynamics}
\author[1]{Grant Rotskoff \thanks{This work was partially supported by the James S. McDonnell Foundation.}}
\author[1,3]{Samy Jelassi}
\author[1,2]{Joan Bruna \thanks{This work was partially supported by the Alfred P. Sloan Foundation and NSF RI-1816753. }}
\author[1]{Eric Vanden-Eijnden \thanks{This work was partially supported by the Materials Research Science and Engineering Center(MRSEC) program of the National Science Foundation(NSF) under award number DMR-1420073 and by NSF under award number DMS-1522767. }}
\affil[1]{Courant Institute of Mathematical Sciences, New York University}
\affil[2]{Center for Data Science, New York University}
\affil[3]{Princeton University}

\begin{document}

\maketitle

\begin{abstract}
  Neural networks with a large number of parameters admit a mean-field
  description, which has recently served as a theoretical explanation
  for the favorable training properties of ``overparameterized'' models.  
  In this regime, gradient descent obeys a
  deterministic partial differential equation (PDE) that converges to
  a globally optimal solution for networks with a single hidden layer
  under appropriate assumptions.  In this work, we propose a non-local
  mass transport dynamics that leads to a modified PDE with the same
  minimizer.  We implement this non-local dynamics as a stochastic
  neuronal birth-death process and we prove that it accelerates the
  rate of convergence in the mean-field limit.  We subsequently
  realize this PDE with two classes of numerical schemes that converge
  to the mean-field equation, each of which can easily be implemented
  for neural networks with finite numbers of parameters.  We
  illustrate our algorithms with two models to provide intuition for
  the mechanism through which convergence is accelerated.
\end{abstract}

\section{Introduction}
As a consequence of the universal approximation theorems, sufficiently wide single layer neural networks are expressive enough to accurately represent a broad class of functions~\cite{cybenko_approximation_1989, barron_universal_1993,park_universal_1991}.
The existence of a neural network function arbitrarily close to a given target function, however, is not a guarantee that any particular optimization procedure can identify the optimal parameters.
Recently, using mathematical tools from optimal transport theory and interacting particle systems, it was shown that gradient descent~\cite{rotskoff_neural_2018, mei_mean_2018, sirignano_mean_2018, chizat_global_2018} and stochastic gradient descent converge asymptotically to the target function in the large data limit.

This analysis relies on taking a ``mean-field'' limit in which the number of parameters $n$ tends to infinity.
In this setting, gradient descent optimization dynamics is described by a partial differential equation (PDE), corresponding to a Wasserstein gradient flow on a convex energy functional.
While this PDE provides a powerful conceptual framework for analyzing the properties of neural networks evolving under gradient descent dynamics, the formula confers few immediate practical advantages.
Nevertheless, analysis of this Wasserstein gradient flow motivates the interesting possibility of altering the dynamics to accelerate convergence.

In this work, we propose a dynamical scheme involving a parameter birth/death process. 
It can be defined on systems of interacting (e.g., neural network optimization) or non-interacting particles. 
We prove that the resulting modified transport equation converges to the global minimum of the loss in both interacting and non-interacting regimes (under appropriate assumptions), and we provide an explicit rate of convergence in the latter case for the mean-field limit. 
Interestingly---and unlike the gradient flow---the \emph{only} fixed point of the dynamics is the global minimum of the loss function.
We study the fluctuations of finite particle dynamics 
around this mean-field convergent solution, showing that they are of the same order throughout the dynamics and therefore providing algorithmic guarantees directly applicable to finite single-layer neural network optimization.
Finally, we derive  algorithms that converge to the birth-death PDEs and verify numerically that these schemes accelerate convergence even for finite numbers of parameters.

Summarily, we describe:

{\bf Global convergence and monotonicity of the energy with birth-death dynamics ---}
We propose in Section \ref{sec:setup} two distinct modifications of the original gradient flow that can be interpreted as birth-death processes. 
In this sense, the processes we describe amount to non-local mass transport in the equation governing the parameter distribution. 
We prove that the schemes we introduce guarantee global convergence and increase the rate of contraction of the energy compared to gradient descent and stochastic gradient descent for fixed $\mu$. We also derive asymptotic rates of convergence (Section \ref{sec:convergence}). 

{\bf Analysis of fluctuations and self-quenching ---} The birth-death dynamics introduces additional fluctuations that are not present in gradient descent dynamics.  In Section \ref{sec:fluctu} we calculate these fluctuations using tools from the theory of measure-valued Markov processes.  We show that these fluctuations, for $n$ sufficiently large, are of order $O(n^{-1/2})$ and ``self-quenching'' in the sense that they diminish in magnitude as the quality as the optimization dynamics approaches the optimum.

{\bf Algorithms for realizing the birth-death schemes ---}
In Section \ref{sec:algo} we detail numerical schemes (and provide implementations in $\texttt{PyTorch}$) of the birth-death schemes described below.
In the particular case of neural networks, the computational cost of implementing our procedure is minimal because no additional gradient computations are required.
We demonstrate the efficacy of these algorithms on simple, illustrative examples in Section \ref{sec:experiments}.

 \section{Related Works}

Non-local update rules appear in various areas of machine learning and optimization. 
Derivative-free optimization \cite{rios2013derivative} offers a general framework for optimizing complex non-convex functions using non-local search heuristics. Some notable examples include Particle Swarm Optimization \cite{kennedy2011particle} and Evolutionary Strategies, such as the Covariance Matrix Adaptation method \cite{hansen2006cma}. These approaches have found some renewed interest in the optimization of neural networks in the context of Reinforcement Learning  \cite{salimans2017evolution,such2017deep} and hyperparameter optimization \cite{jaderberg2017population}.

Our setup of non-interacting potentials is closely related to the so-called Estimation of Distribution Algorithms \cite{baluja1995removing, larranaga2001estimation}, which define update rules for a probability distribution over a search space by querying the values of a given function to be optimized. In particular, Information Geometric Optimization Algorithms \cite{ollivier2017information} study the dynamics of parametric densities using ordinary differential equations, focusing on invariance properties. In contrast, our focus in on the combination of transport (gradient-based) and birth/death dynamics.  

Dropout \cite{srivastava2014dropout} is a regularization technique popularized by the AlexNet CNN \cite{krizhevsky2012imagenet} reminiscent of a birth/death process, but we note that its mechanism is very different: rather than killing a neuron and replacing it by a new one with some rate, Dropout momentarily masks neurons, which become active again at the same position; in other words, Dropout  implements a purely local transport scheme, as opposed to our non-local dynamics. 

Finally, closest to our motivation is \cite{wei_margin_2018}, who, building on the recent body of works that leverage optimal transport techniques to study optimization in the large parameter limit \cite{rotskoff_neural_2018,chizat_global_2018,mei_mean_2018,sirignano_mean_2018}, proposed a modification of the dynamics that replaced traditional stochastic noise by a resampling of a fraction of neurons from a base, fixed measure. Our model has significant differences to this scheme, namely we show that the dynamics preserves the same global minimizers and accelerates the rate of convergence. Finally, our interpretation of the modified dynamics in terms of a generalized gradient flow is related to the unbalanced optional transport setups of \cite{kondratyev2016,Liero:2018bz,CHIZAT20183090}. 
 \section{Mean-field PDE and Birth-death Dynamics}
\label{sec:setup}

\subsection{Mean-Field Limit and Liouville dynamics}
Gradient descent propagates the parameters locally in proportion to the gradient of the objective function.
In some cases, an optimization algorithm can benefit from nonlocal dynamics, for example, by allowing new parameters to appear at favorable values and existing parameters to be removed if they diminish the quality of the representation.
In order to exploit a nonlocal dynamical scheme, it is useful to interpret the parameters as a system of $n$ particles, $\thetab_i\in D$, a $k$-dimensional differentiable manifold, which for $i=1,\ldots,n$ evolve on a landscape  determined by the objective function $ \ell(\thetab_1, \dots, \thetab_n)$.  Here we will focus on situations where the objective function may involve interactions between pairs of parameters:
\begin{equation}
\label{eq:interactell}
\ell(\thetab_1,\dots,\thetab_n) = \sum_{i=1}^n F(\thetab_i)
+ \frac{1}{2n}\sum_{i,j=1}^n K(\thetab_i, \thetab_j)
\end{equation}
where $F:D\to \RR$ is a single particle energy function and $K:D\times D \to \RR$ is a symmetric semi-positive definite interaction kernel.
Interestingly, optimizing neural networks with the mean-squared loss function fits precisely this framework \cite{rotskoff_neural_2018, mei_mean_2018, chizat_global_2018}.
Consider a supervised learning problem using a neural network with nonlinearity $\varphi$.
If we write the neural network as
\begin{equation}
    \label{eq:NN}
    f_n(\xb;\thetab_1,\dots,\thetab_n) = \frac1n\sum_{i=1}^n \varphi(\xb, \thetab_i)
\end{equation}
and expand the loss function, 
\begin{equation}
    \ell(\thetab_1,\dots,\thetab_n) = \tfrac12 \EE_{y,\xb} \left|y - f_n(\xb; \thetab_1,\dots,\thetab_n)\right|^2,
\end{equation}
we see that, up to an irrelevant constant  depending only on the data distribution, we arrive at \eqref{eq:interactell} with
\begin{equation}
  \label{fefe}
    F(\thetab) = -\EE_{y,\xb} \bigl[ y \varphi(\xb, \thetab) \bigr],
\end{equation}
and,
\begin{equation}
\label{keke}
K(\thetab, \thetab') =
\EE_{\xb} \bigl[\varphi(\xb, \thetab) \varphi(\xb, \thetab')\bigr].
\end{equation} 
We also consider \emph{non-interacting} objective functions in which $K=0$ in~\eqref{eq:interactell}. 
Optimization problems that fit this framework include resource allocation tasks in which, e.g., weak performers are eliminated, Evolution Strategies, and Information Geometric Optimization \cite{ollivier2017information}.

In the case of gradient descent dynamics, the evolution of the particles $\thetab_i$ is governed for $i=1,\ldots,n$ by
\begin{equation}
  \label{eq:GD}
  \dot{\thetab_i} = - \grad_{\thetab_i}  \ell(\thetab_1, \dots, \thetab_n).
\end{equation}
To analyze the dynamics of this particle system, we consider the ``mean-field'' limit $n\to \infty$.
As the number of particles becomes large, the empirical distribution of particles
\begin{equation}
\mu^{(n)}_t(d\thetab) =\frac1n\sum_{j=1}^n \delta_{\thetab_j(t)}(d\thetab)
\label{eq:rhon}
\end{equation}
leads to a deterministic partial differential equation at first order~\cite{rotskoff_neural_2018, mei_mean_2018, chizat_global_2018, sirignano_mean_2018},
\begin{equation}
  \partial_t \mu_t = \div \left( \mu_t \grad V \right),
  \label{eq:pde}
\end{equation}
where $\mu_t$ is the weak limit of $\mu^n_t$ and $\mu_0$ is some distribution from which the initial particle positions $\thetab_i(0)$ are drawn independently.
The potential $V:D \to \RR$ is specified by the objective function $ \ell$ as
\begin{equation}
    \label{eq:potential}
    V(\thetab, [\mu]) = F(\thetab) + \int_D K(\thetab,\thetab') \mu(d\thetab').
\end{equation}
and \eqref{eq:pde} should be interpreted in the weak sense in general:
\begin{equation}
  \forall \phi \in C^\infty_c(D) \ : \qquad \partial_t \int_D \phi(\thetab) \mu_t(d\thetab)  = - \int_D \nabla \phi(\thetab)\cdot  \grad V (\thetab, [\mu_t]) \mu_t(d\thetab),
  \label{eq:pdeweak}
\end{equation}
where $C^\infty_c(D)$ denotes the space of smooth functions with compact support on $D$.

Interestingly, $V$ is the gradient with respect to $\mu$ of an energy functional $\mathcal{E}[\mu]$, 
\begin{equation}
    \label{eq:energy}
    \mathcal{E}[\mu] = \int_D F(\thetab) \mu(d\thetab)+ \tfrac12 \int_{D\times D} K(\thetab,\thetab') \mu(d\thetab)\mu(d\thetab').
\end{equation}
As a result, 
the nonlinear Liouville equation~\eqref{eq:pde} is the Wasserstein gradient flow with respect to the energy functional $\mathcal{E}[\mu]$.
Local minima of $V$ (where $\grad V=0$) are clearly fixed points of this gradient flow, but these fixed points may not always be minimizers of the energy when $\supp \mu \subset D$.
When the initial distribution of parameters has full support, neural networks evolving with gradient descent avoid these spurious fixed points under appropriate assumptions about their nonlinearity~\cite{chizat_global_2018, rotskoff_neural_2018,mei_mean_2018}.

\subsection{Birth-Death augmented Dynamics}

Here we consider a more general dynamical scheme that involves nonlocal transport of particle mass. As we shall see in Section \ref{sec:convergence}, this dynamics avoids spurious fixed points and local minima, and converges asymptotically to the global minimum.
Consider the following modification of the Wasserstein gradient flow above:
\begin{equation}
  \partial_t \mu_t = \div \left( \mu_t \grad V \right) - \alpha V \mu_t \qquad (\alpha>0).
  \label{eq:pde_bd}
\end{equation}
The additional term $- \alpha V \mu_t$ is a birth/death term that
modifies the mass of $\mu$.  If $V$ is positive, this mass will
decrease, corresponding to the removal or ``death'' of parameters.  If
$V$ is negative, this mass will increase, which can be implemented as
duplication or ``cloning'' of parameters.  For a finite number of
parameters, this dynamics could lead to changes in the architecture of
the network.  In many applications it is preferable to fix the total
population, achieved by simply adding a conservation term to the
dynamics,
\begin{equation}
  \partial_t \mu_t = \div \left( \mu_t \grad V \right) - \alpha V \mu_t + \alpha \bar V  \mu_t,
  \label{eq:pde_control}
\end{equation}
where $\bar V\equiv \int_D V d\mu_t$. This equation (like~\eqref{eq:pde_bd}) should in general be interpreted in the weak sense. Here we will focus on solutions of \eqref{eq:pde_control} for the initial condition $\mu_0 \in \mathcal{M}(D)$,  the space of probability measures on~$D$, that satisfy
\begin{equation}
  \label{eq:15}
  \int_D \phi(\thetab) \mu_t(d\thetab) = C^{-1}(t) \int_D
  \phi(\Thetab(t,\thetab)) e^{-\alpha\int_0^t V(\Thetab(s,\thetab),\mu_s]) ds} \mu_0(d\thetab)
\end{equation}
where $\phi:D\to\RR$ is any bounded differentiable function with bounded gradient, $C(t)$ is given by
\begin{equation}
  \label{eq:18}
  C(t) = e^{-\alpha \int_0^t \bar V[\mu_s] ds} \equiv \int_D e^{-\alpha \int_0^t V(\Thetab(s,\thetab),[\mu_s]) ds} \mu_0(d\thetab),
\end{equation}
and $\Thetab(t,\thetab)$ satisfies
\begin{equation}
  \label{eq:16}
  \dot \Thetab(t,\thetab) = - \nabla V(\Thetab(t,\thetab),[\mu_t]), \qquad
  \Thetab(0,\thetab) = \thetab.
\end{equation}
Formula~\eqref{eq:15} can be formally established by solving~\eqref{eq:pde_control} by the method of characteristics. In the non-interacting case, since $V(\thetab,[\mu_t]) = F(\thetab)$, \eqref{eq:15} is explicit and well-posed under appropriate assumptions on $F$ (see Assumption~\ref{as:noninter} below). In the interacting case, \eqref{eq:15} is implicit since the right hand side depends on $\mu_t$.  Following Chizat \& Bach~\cite{chizat_global_2018}, we know that under appropriate assumptions on $F$ and $K$ (see Assumption~\ref{as:interacting-case} below), solutions to  \eqref{eq:15} exist for all $t>0$ for appropriate initial $\mu_0$ that are compactly supported in $D$. Here we will assume global existence of solutions to this equation for $\mu_0$ such that $\supp \mu_0 = D$ with $D$ open: if $\mu_0$ decays sufficiently fast at infinity, this assumption is supported by the alternative derivation of~\eqref{eq:pde_bd} based on a proximal gradient formulation given in Sec.~\ref{sec:proxi}.

Note that solutions of \eqref{eq:pde_bd} that satisfy \eqref{eq:15} are probability measures since they are positive by definition and  we can set $\phi=1$ in~\eqref{eq:15} to deduce that $\mu_t(D)=1$. We can also show that the birth-death terms improve the rate of energy decay, as stated in the following proposition:
\begin{proposition}
\label{prop:consis1}
Let $\mu_t$ be a solution of (\ref{eq:pde_control}) for the initial condition $\mu_0 \in \mathcal{M}(D)$ that satisfies \eqref{eq:15} for all $t\ge0$. 
Then, $\mu_t(D)=1$ for all $t\ge0$, and
$E(t)=\mathcal{E}[\mu(t)]$ satisfies 
\begin{equation}
 \label{eq:Edecay}
    \dot E(t) = - \int_D | \nabla V(\thetab,[\mu_t])|^2 \mu_t(d\thetab)
    - \alpha\int_D \left(V(\thetab,[\mu_t])-\bar V[\mu_t] \right)^2 \mu_t(d\thetab) \le 0.
\end{equation}
\end{proposition}
\noindent
\textit{Proof:} \eqref{eq:Edecay} can be formally obtained by testing  (\ref{eq:pde_control}) against $V(\thetab,[\mu_t])$ and using the chain rule to deduce that $d\mathcal{E}[\mu_t]/dt = \int_D V(\thetab,[\mu_t]) \partial_t \mu_t(d\thetab)$. To complete the proof, we need to show that this testing is legitimate and the terms at the right hand side of \eqref{eq:Edecay} are well-defined; this is done in Appendix~\ref{app:proof_global_convergence} by differentiating $C(t)$. \hfill $\square$

The birth-death term thus contributes to increase the rate of decay of the energy at all times. A natural question is whether such improved energy decay can lead to global convergence of the dynamics to the global minimum of the energy.
As it turns out, the answer is yes: the fixed points of the
birth-death PDEs~\eqref{eq:pde_bd} and~\eqref{eq:pde_control} are the
global minimizers of the energy $\mathcal{E}[\mu]$, as we prove in
Section \ref{sec:convergence}. How to implement a particle dynamics
consistent with~\eqref{eq:pde_control} is discussed in Sections
\ref{sec:fluctu} and \ref{sec:algo}.

We also note that there are several ways in which we can
modify~(\ref{eq:pde_control}) to certain advantages: this is discussed
in Appendix~\ref{sec:modify}.

\subsection{Proximal formulation of birth-death dynamics}
\label{sec:proxi}

Following the frame of Ref.~\cite{jordan1998variational}, we can give an alternative interpretation to the
birth-death PDE~\eqref{eq:pde_control}. 
First, we recall that the
PDE~\eqref{eq:pde} can be obtained as the time-continuous limit
$(\tau \to0)$ of the proximal optimization scheme (also known as minimizing movement scheme~\cite{Santambrogio:2017jr}) in which a sequence
of distributions $\{\mu_k\}_{k\in \NN_0}$ is constructed via the
iteration: given an initial $\mu_0$ such that $\mathcal{E}[\mu_0]<\infty$, set
\begin{equation}
  \label{eq:proxi1}
  \mu_{k+1} \in \argmin \left(\mathcal{E}[\mu] + \tfrac12\tau^{-1} W_2^2(\mu,\mu_k)\right), \qquad k=0,1,2,\ldots,
\end{equation}
where $W_2(\mu,\mu_k)$ denotes the $2$-Wasserstein distance between
the probability measures $\mu$ and $\mu_k$.  
Interestingly, the birth-death PDE relies on a different measure of ``distance'': the PDE
\begin{equation}
  \label{eq:proxi1a}
  \partial_t \mu_t = - \alpha V\mu_t+\alpha \bar V \mu_t,
\end{equation}
can be obtained as the time-continuous limit of the proximal
optimization scheme: given an initial $\mu_0$ such that $\mathcal{E}[\mu_0]<\infty$, set
\begin{equation}
  \label{eq:proxi2}
  \mu_{k+1} \in \argmin \left(\mathcal{E}[\mu] + (\alpha\tau)^{-1} D_{\text{KL}}(\mu
  || \mu_k)\right), \qquad k=0,1,2,\ldots,
\end{equation}
where the minimum is taken over all probability measures
$\mu \in \mathcal{M}(D)$ and $D_{\text{KL}}(\mu || \mu_k)$ is the
Kullback-Leibler divergence
\begin{equation}
  \label{eq:proxi3}
  D_{\text{KL}}(\mu
  || \mu_k) = \int_D \log\left( \frac{d\mu}{d\mu_k}\right) d\mu~.
\end{equation}
We verify this claim formally; notice that the Euler-Lagrange equation
for the minimizer $\mu_{k+1}$, obtained by zeroing the first variation
of the objective function in~\eqref{eq:proxi2}, reads
\begin{equation}
  \label{eq:proxi4}
  V(\thetab,[\mu_{k+1}]) + (\alpha\tau)^{-1} \log\left(
    \frac{d\mu_{k+1}}{d\mu_k}\right) + \lambda= 0
\end{equation}
where $\lambda$ is a Lagrange multiplier added to enforce $\int_D
d\mu_{k+1}=1$. \eqref{eq:proxi4} can be reorganized into
\begin{equation}
  \label{eq:proxi5}
  \mu_{k+1} = C^{-1} \mu_k \exp\left(-\alpha\tau V(\thetab,[\mu_{k+1}]) \right)
\end{equation}
where $C$ is adjusted so that $\int_D
d\mu_{k+1}=1$. \eqref{eq:proxi5} is the discrete equivalent of~\eqref{eq:15} If $\tau$ is small, we can expand the exponential to
arrive at 
\begin{equation}
  \label{eq:proxi6}
  \mu_{k+1} = C^{-1} \left( \mu_k -\alpha\tau V(\thetab,[\mu_{k+1}])\mu_k + O(\tau^2)\right)
\end{equation}
Setting $\mu_{k+1} = \mu_{k} + O(\tau)$ in $V$ and expanding again
gives
\begin{equation}
  \label{eq:proxi7}
  \mu_{k+1} = \mu_k -\alpha\tau V(\thetab,[\mu_{k}]) \mu_k + \alpha\tau
  \left(\int_D V(\thetab,[\mu_{k}]) d\mu_k \right) \mu_k +O(\tau^2) 
\end{equation}
where we have also expanded $C$ and solved for it explicitly at
leading order in $\tau$. Subtracting $\mu_k$ for both sides, dividing
by $\tau$, and letting $\tau\to0$ gives~\eqref{eq:proxi1a}. The full
PDE~(\ref{eq:pde_control}) can be obtained by
alternating~(\ref{eq:proxi1}) and~(\ref{eq:proxi2}). 

Note that, under Assumption~\ref{as:interacting-case} below, the energy $\mathcal{E}[\mu]$ is convex and bounded below. As a result the augmented functionals to minimize in both~\eqref{eq:proxi1} and \eqref{eq:proxi2} are strictly convex, which means that they admit a unique minimizer. This shows that the  measures in the sequence $\{\mu_k\}_{k\in\NN_0}$ are well-defined and such that $\mathcal{E}[\mu_{k+1}]\le \mathcal{E}[\mu_{k}]$ whether we use~\eqref{eq:proxi1}, \eqref{eq:proxi2}, or alternate between both.  Because we discretize time in practice, solutions of~\eqref{eq:pde_control} satisfying \eqref{eq:15} for all $t>0$ can be interpreted as implementations of the proximal scheme. 
Taking the limit $\tau \to 0$ with $k\tau$ large, however, requires ensuring well-definedness of the terms on the right hand side of~\eqref{eq:pde_control}.   
This proximal interpretation also enables the design of distinct algorithms for implementing
this PDE at particle level.

 \section{Convergence of Transport Dynamics with Birth-death}
\label{sec:convergence}

Here, we compare the solutions of the original PDE~\eqref{eq:pde} with
those of the PDE~\eqref{eq:pde_control} with birth-death. We restrict
ourselves to situations where $F$ and $K$ in~\eqref{eq:energy} are
such that $\mathcal{E}[\mu]$ is bounded from below.  Our main
technical contributions are results about convergence towards global
energy minimizer as well as convergence rates as the
dynamics approaches these minimizers. 
We consider separately the non-interacting and the interacting cases.

Under gradient descent dynamics, global convergence can be established
with appropriate assumptions on the initialization and architecture of
the neural network.  \cite{mei_mean_2018} establishes global
convergence and provides a rate for neural networks with bounded
activation functions evolving under stochastic gradient descent.
Similar results were obtained
in~\cite{chizat_global_2018,rotskoff_neural_2018}, in which it is
proven that gradient descent converges to the globally optimal
solution for neural networks with particular homogeneity conditions on
the activation functions and regularizers.  Closely related to the
present work, \cite{wei_margin_2018} provides a convergence rate for a
``perturbed'' gradient flow in which uniform noise is added to the
PDE~\eqref{eq:pde}.  It should be emphasized that, unlike our
formulation, the addition of uniform noise changes the fixed point of
the PDE and convergence to only an approximate global solution can be
obtained in that setting.

\subsection{Non-interacting Case}
\label{sec:noninteracting}

We consider first the non-interacting case with $V=F$ and $D=\RR^k$,
under

\begin{assumption}
  \label{as:noninter}
  $F\in C^2(\RR^k)$ is a Morse function, coercive, and with a single
  global minimum located at~$\thetab^*$.
\end{assumption}
With no loss of generality we set $F(\thetab^*) =0$ since adding an
offset to $F$ in (\ref{eq:pde_control}) does not affect the dynamics.
We also denote by $H^*=\nabla\nabla F(\thetab^*)$ the Hessian of $F$
at $\thetab^*$: recall that a Morse function is such that its Hessian
is nondegenerate at all its critical points (where $\nabla F = 0$) and
it is coercive if $\lim_{\thetab\to\infty} F(\thetab) = \infty$. Our
main result is
\begin{theorem}[Global Convergence and Rate: Non-interacting Case]
\label{th:local-non-rate}
Assume that the initial condition $\mu_0$ of the PDE~\eqref{eq:pde_bd}
has a density $\rho_0$ positive everywhere in $\RR^k$ and is such that
$\mathcal{E}[\mu_0] < \infty$. Then under Assumption~\ref{as:noninter}
the solution of~\eqref{eq:pde_bd} satisfies
\begin{equation}
    \mu_t \rightharpoonup \delta_{\thetab^*} \qquad \text{as \ \ $t\to \infty$.}
  \end{equation}
  In addition we can quantify the convergence rate: if
  $\bar F(t) = \int_{\RR^k} F(\thetab)\mu_t(\thetab)$, then
  $\exists C>0$ such that $\forall\epsilon>0$, the time $t_\epsilon$
  needed to reach $\mathcal{E}[\mu_{t_\epsilon}] \leq \epsilon$
  satisfies
\begin{equation}
\label{lu00}
t_\epsilon \leq C \epsilon^{-(d+2)/2}.
\end{equation}
Furthermore the rate of convergence becomes exponential in time
asymptotically:  for all $\delta >0$, $\exists t_\delta$ such that
\begin{equation}
    \label{eq:expdecay}
    \bar F(t)\le \alpha^{-1} \tr \left(H^* e^{-2 H^* (t-\delta)}\right)
    \quad
    \text{if \ \ $t\ge t_\delta$}.
\end{equation}
\end{theorem}
In fact we show that
\begin{equation}
  \label{eq:liminonint}
  \lim_{t\to\infty} \frac{\alpha \bar F(t) }{\tr \left(H^* e^{-2 H^*
        t}\right) } =1.
\end{equation}
The theorem is proven in Appendix~\ref{sec:convergence-noninteracting}
This proof shows that the additional birth-death terms in the
PDE~\eqref{eq:pde_bd} allow the measure to concentrate rapidly in the
vicinity of~$\thetab^*$; subsequently, the transport term takes over
and leads to the exponential rate of energy decay
in~\eqref{eq:expdecay}.  The proof also shows that, if we remove the
transportation term $\div \left( \mu_t \grad V \right)$ in the
PDE~\eqref{eq:pde_bd}, the energy only decreases linearly in time
asymptotically.  This means that the combination of the transportation
and the birth-death terms accelerates convergence.  A similar theorem
can be proven for the PDE~\eqref{eq:pde_prior}.

\subsection{Interacting Case}

Let us now consider the interacting case, when $V$ is given
by~\eqref{eq:potential} with $K\not=0$. We make

\begin{assumption}
  \label{as:D}
  The set $D$ is a $k$-dimensional differentiable manifold which is
  either closed (i.e. compact, with no boundaries), or open (i.e. with
  no closed subset), or the Cartesian product of a closed and an open
  manifold.
\end{assumption}
\begin{assumption}
  \label{as:interacting-case}
  The kernel $K$ is symmetric, positive semi-definite, and twice
  differentiable in its arguments, $K\in C^2(D\times D)$; 
  $F\in C^2(D)$; and $F$ and $K$ are such that the energy is bounded
  from below, i.e. $\exists m\in \RR$ such that $\forall \mu \in
  \mathcal{M}(D)$ : $\mathcal{E}[\mu]\ge m$.
\end{assumption}
This technical assumption typically holds for neural networks.
Assumption~\ref{as:interacting-case} guarantees that the quadratic
energy $\mathcal{E}[\mu]$ in~\eqref{eq:energy} has a (unique) minimum
value.  While we cannot guarantee in general that this minimum is
reached only by minimizers, below we will work under the assumption
that minimizers exist. These are solutions in~$\mathcal{M}(D)$ of
following Euler-Lagrange equations:
\begin{equation}\label{eq:euler_lagrange}
  \left\{
  \begin{aligned}
    V(\thetab, [\mu_*]) &= \bar V[\mu_*] \qquad &&\forall \thetab \in \supp \mu_*\\
    V(\thetab, [\mu_*]) &\ge \bar V[\mu_*] \qquad &&\forall \thetab
    \in D.
  \end{aligned}\right.
\end{equation}
where $\bar V[\mu] \equiv \int_D V(\thetab,[\mu])
\mu(d\thetab)$. These equations are
well-known~\cite{serfaty_coulomb_2015}: for the reader's convenience we
recall their derivation in Appendix~\ref{sec:euler-lagrange}.

Minimizers of the energy should not be confused with fixed points of
the dynamics.  In particular, a well-known issue with the
PDE~\eqref{eq:pde} is that it potentially has many more fixed points
than $\mathcal{E}[\mu]$ has minimizers: Indeed, rather
than~(\ref{eq:euler_lagrange}), these fixed points only need to
satisfy
\begin{equation}
    \label{eq:statiopoints}
    \nabla V(\thetab,[\mu]) = 0 \qquad \forall \thetab \in \supp \mu.
\end{equation}
It is therefore remarkable that, if we pick an initial condition
$\mu_0$ for the birth-death PDE~\eqref{eq:pde_control} that has full
support, the solution to this equation converges to a global minimizer
of $\mathcal{E}[\mu]$:
\begin{theorem}[Global Convergence to Global Minimizers: Interacting Case]
\label{th:global}
Let $\mu_t$ denote the solution of~\eqref{eq:pde_control} that satisfies~\eqref{eq:15} for the
initial condition $\mu_0$ with $\supp \mu_0 = D$.  If
$\mu_t\rightharpoonup \mu_*$ as $t\to\infty$ for some probability
measure $\mu_*\in \mathcal{M}(D)$, then under Assumptions~\ref{as:D}
and \ref{as:interacting-case} $\mu_*$ is a global minimizer of
$\mathcal{E}[\mu]$.
\end{theorem}
This theorem is proven in
Appendix~\ref{app:proof_global_convergence}. Note that the theorem
holds under the assumption that $\mu_t$ converges to a fixed point
$\mu_*$, which we cannot guarantee \textit{a~priori} but should be
true for a wide class of $F$ and $K$ and initial conditions $\mu_0$
satisfying properties like $\mathcal{E}[\mu_0],\infty$---for more
details on these conditions see the proof in
Appendix~\ref{app:proof_global_convergence}.  One aspect of this proof
is based on the evolution equation~\eqref{eq:Edecay} for
$\mathcal{E}[\mu_t]$.  Since $d \mathcal{E}[\mu_t]/dt \le0$ and since
$\mathcal{E}[\mu_t]$ is bounded from below by
Assumption~\ref{as:interacting-case}, by the bounded convergence
theorem, the evolution must stop eventually. By assumption, this
involves $\mu_t$ converging weakly towards some $\mu_*$. This happens
when both integrals in~\eqref{eq:Edecay} are zero, i.e. $\mu_*$ must
satisfy the first equation in~\eqref{eq:euler_lagrange} as well
as~\eqref{eq:statiopoints}. What remains to be shown is that $\mu_*$
must also satisfy the second equation in~\eqref{eq:euler_lagrange},
which we check in Appendix~\ref{app:proof_global_convergence}.

Regarding the rate of convergence, we have the following result:

\begin{theorem}[Asymptotic Convergence Rate: Interacting Case]
  \label{th:local-int-rate} Under the same conditions as in
  Theorem~\ref{th:global}, $\exists C>0$ and $t_C>0$ such that
  $E(t) = \mathcal{E}[\mu_t] -\mathcal{E}[\mu_*]\ge 0$ satisfies
\begin{equation}
  E(t) \le C t^{-1}
  \quad \text{if \ \ $t\ge t_C$}
\end{equation}
\end{theorem}
The proof of this theorem is given in
Appendix~\ref{sec:convergence-interacting}  where we show that
\begin{equation}
  \label{eq:27}
  \lim_{t\to\infty} t E(t) \le C \in (0,\infty].
\end{equation}

 \section{From Mean-field to Particle Dynamics with Birth-Death }
\label{sec:fluctu}

In practice the number of parameters $n$ is finite, so we must verify
that we can implement dynamics at finite particle numbers that is
consistent with the PDEs with birth-death terms introduced in
Sec.~\ref{sec:setup} in the mean-field limit $n\to \infty$. We must
also ensure that the fluctuations arising from the discrete particles
do not pose a problem for the optimization dynamics. In this section,
we carry out this program in the context of the
PDE~\eqref{eq:pde_control}. Analogous calculations can be performed in
the case of~\eqref{eq:pde_prior}. These results rely on the theory of
measure-valued Markov processes~\cite{dawson2006measure}, and are
detailed in Appendix~\ref{app:flucts}.

The dynamics of the particles $\{\thetab_i(t)\}_{i=1}^n$ is specified
by a Markov process defined as follows: the birth-death part of the
evolution is realized by equipping each particle $\thetab_i$ with an
independent exponential clock with (signed) rate
\begin{equation}
  \tilde V(\thetab_i) = F(\thetab_i)+\frac1n
  \sum_{j=1}^nK(\thetab_i,\thetab_j) -\frac1n \sum_{j=1}^n \left(F(\thetab_j) +\frac1n
  \sum_{k=1}^nK(\thetab_j,\thetab_k)\right)
  \label{eq:centeredrateinteract}
\end{equation}
such that:
\begin{enumerate}
\item If $\tilde V(\thetab_i(t)) >0$, the particle $\thetab_i$ is
  duplicated with instantaneous rate $\alpha \tilde V(\thetab_i(t))$,
  and a particle $\thetab_j$ chosen at random in the stack is killed
  to preserve the population size.
\item If $\tilde V(\thetab_i(t)) < 0$, the particle $\thetab_i$ is
  killed with instantaneous rate $ \alpha |\tilde V(\thetab_i(t))|$,
  and a particle $\thetab_j$ chosen at random in the stack is
  duplicated to preserve the population size.
\end{enumerate}
Between these birth events the particles evolve by the GD flow
\eqref{eq:GD}.

Due to the interchangeability of the particles, the evolution of their
empirical distribution $\mu^{(n)}_t$ defined in~\eqref{eq:rhon} is
also Markovian: it is referred to in the probability literature as a
\textit{measured-valued Markov process}~\cite{dawson2006measure}. We
can write down the generator of this process, which specifies the
evolution of the expectation of functionals of $\mu^{(n)}_t$, and
analyze its behavior as $n\to\infty$. These calculations are performed
in Appendix~\ref{app:flucts}, and they lead to:
\begin{proposition}[Law of Large Numbers]
  \label{th:lln}
  Let the empirical distribution of the initial position of the
  particles be $\mu_0^{(n)}= n^{-1}\sum_{i=1}^n \delta_{\thetab_i(0)}$
  and assume that $\mu^{(n)}_0 \rightharpoonup \mu_0$ as $n\to\infty$.
  Then, for all for $t\in [0,\infty)$,
  $\mu_t^{(n)}= n^{-1}\sum_{i=1}^n \delta_{\thetab_i(t)}
  \rightharpoonup \mu_t$ in law as $n\to\infty$, where $\mu_t$
  satisfies \eqref{eq:pde_control} with the initial condition $\mu_{t=0} = \mu_0$.
\end{proposition} 
This statement verifies that, to leading order, the large particle
limit recovers the mean-field PDE~\eqref{eq:pde_control}.

While the limit gives rise to the birth-death term of the PDE as
expected, we can also quantify the scale and asymptotic behavior of
the higher order fluctuations at finite $n$. This computation ensures
that finite $n$ fluctuations do not overcome the convergence expected
from the mean-field analysis. To do so, we we introduce the
discrepancy distribution defined by the difference, scaled by
$\sqrt{n}$, between the empirical distribution and its mean-field
limit
\begin{equation}
  \label{eq:discrepencydistribution}
   \omega^{(n)}_t \equiv \sqrt{n} \left(\mu^{(n)}_t - \mu_t \right)
 \end{equation}
 where $\mu^{(n)}_t$ is the empirical distribution defined
 in~(\ref{eq:rhon}) and $\mu_t$ is limit
 satisfying~\eqref{eq:pde_control2}. We can then analyze the generator
 of the joint process $(\mu_t, \omega^{(n)}_t )$ and deduce the
 following proposition:
 \begin{proposition}[Central Limit Theorem]
   \label{th:clt}
  In the limit as $n\to\infty$, we have
\begin{equation}
  \omega^{(n)}_t 
  \rightharpoonup \omega_t \qquad \textrm{in law}
\end{equation}
where $\omega_t$ is Gaussian random distribution with zero mean and
whose covariance satisfies a linear equation with a source term
proportional to $\alpha|\tilde V(\thetab,[\mu_t])|\mu_t$, see
\eqref{eq:covarianceinter} in Appendix~\ref{app:flucts}.
\end{proposition}

The key consequence of this proposition is that it specifies the scale
of the fluctuations of $\mu^{(n)}_t$ above its mean field limit
$\mu_t$. First it shows that these fluctuations are on a scale
$O(\sqrt{\alpha/n})$. This is why $\alpha$ should be kept $O(1)$
relative to $n$. While it may appear that increasing $\alpha$
accelerates the rate of convergence at mean-field level, the
fluctuations would grow and the $n\to \infty$ and $\alpha \to \infty$
limit do not commute. Second, the relation between the scale of the noise
and the magnitude of $|\tilde{V}|\mu_t$ has an important consequence
for the convergence of the dynamics: because $|\tilde{V}|\mu_t\to 0$
as $t\to\infty$, the fluctuations are ``self-quenching'' in the sense
that their amplitude diminishes and eventually vanishes as
$\mu_t \to \mu_*$. In particular, for both the interacting and
non-interacting cases, the only stable fixed point of the equation for
the covariance of $\omega_t$ is zero.

We should emphasize that these conclusions rely on $n$ being large
enough that both the LLN and the CLT apply. In practical situations,
it may be difficult to determine the threshold value of $n$ to reach
this regime---it may grow with the dimension of $D$. At finite~$n$, we
also cannot rule out the possibility of some distinct dynamical regime
in which the fluctuations grow with time---our results simply indicate
that, in the regime where the LLN and CLT apply, the timescale for
such a phenomenon would be diverging with $n$. These concerns are
partially placated by the fact that our experiments show no signs of
any such distinct dynamical regime and clearly indicate that
birth-death helps accelerating convergence at moderate values of~$n$.

Finally we want to stress that, while the calculations above indicate
convergence with the birth-death dynamics alone when $n$ is large
enough, the gradient flow probably plays a crucial part in
accelerating the underlying optimization procedure, especially at
moderate values of $n$.  Without the transport term, the birth-death
dynamics can only adjust the weight of existing neurons, which is
clearly inefficient in some cases. That is, we do not advocate the use of
birth-death dynamics alone, but rather to combine it with GD.

 \section{Algorithms}
\label{sec:algo}

Numerical schemes that converge to the PDEs presented in
Sec.~\ref{sec:setup} are both straightforward to design and easy to
implement. In absence of the GD part of the dynamics, we could use
Kinetic Monte Carlo (also called the Gillespie algorithm) to simulate
birth-death without time-discretization error. However, in the large parameter regime,
this would be computationally expensive: every particle has its own exponential clock,
and the time between successive birth-death events scales like
$1/n$. 
Because we must time-discretize the GD flow, we carry out the birth-death dynamics using the
same time-discretization. 

Denote by $\{ \thetab_i \}_{i=1}^n$ the current configuration of $n$
particles in the interacting potential $\ell$
in~\eqref{eq:interactell}. To update the state of these particles, we
first consider the effect of the GD flow alone, using a
time-discretized approximation of this flow with step of size $\Delta t>0$. With
the forward Euler scheme, this amounts to updating the particle
positions as
\begin{equation}
  \thetab_i \leftarrow \thetab_i - \nabla
  F(\thetab_i) \Delta t
  -\frac1n \sum_{j=1}^n \nabla K(\thetab_i,
  \thetab_j) \Delta t
\end{equation}
While this type of update is standard in machine learning, more accurate integration schemes could be used. 

To implement the birth-death part of the dynamics, we calculate
the probability of survival of the particles assuming that their
position was fixed at the current values $\{ \thetab_i \}_{i=1}^n$
using the empirical value $\tilde V(\thetab_i)$ given in
\eqref{eq:centeredrateinteract} for the rate $V-\bar V$. If $\tilde V(\thetab_i)>0$
the probability that particle $\thetab_i$ be killed in the time
interval of size $\Delta t$ is 
\begin{equation}
  \label{eq:probkilled}
  1 - \exp(\tilde V(\thetab_i) \Delta t)
\end{equation}
Similarly, the probability that it is duplicated in that time interval
if $\tilde V(\thetab_i)<0$ is
\begin{equation}
  \label{eq:probdupli}
  1 - \exp(|\tilde V(\thetab_i) |\Delta t)
\end{equation}
Particles are killed and duplicated in a loop according to this
rule. Since $\sum_{i=1}^n \tilde V(\thetab_i) = 0$ by construction,
this operation preserves the number of particles on average. To
enforce strict population control, we add an additional loop that
guarantees the total population remains fixed after the dynamics above.  
The details are given in Algorithm~\ref{alg:scheme1}. 

The corresponding particle system is a discretized version, both in
particle number and time, of the PDE~\eqref{eq:pde_control} and it
converges to this equation as $n\to\infty$ and $\Delta t \to0$. The
error we make at finite $n$ is analyzed in Sec.~\ref{sec:fluctu}; the
error we make at finite $\Delta t$ can be deduced from standard
results about time discretization of differential equations: with the
Euler scheme used above, this error scales as $O(\Delta t)$.

\begin{algorithm}
  \caption{Parameter birth-death dynamics consistent with~\eqref{eq:pde_control} }\label{alg:scheme1}
\begin{algorithmic}
  \State $\Delta t$, initial $\{ \thetab_i \}_{i=1}^n$ given
  \State $\epsilon = \epsilon_{\rm tol}$, the tolerance \While
  {$\epsilon\geq \epsilon_{\rm tol}$} \For{$i=1:n$} \State set
  $\thetab_i \leftarrow \thetab_i - \nabla F(\thetab_i) \Delta t
  -\frac1n \sum_{j=1}^n \nabla K(\thetab_i, \thetab_j) \Delta t$
  \State calculate $\tilde V(\thetab_i) = F(\thetab_i) +
  n^{-1}\sum_{j=1}^n K(\thetab_i,\thetab_j)- n^{-1} \sum_{j=1}^n \left( F(\thetab_j) + n^{-1}\sum_{k=1}^n K(\thetab_j,\thetab_k)\right)$ 
  \If { $\tilde V(\thetab_i) > 0$  }  \State { kill $\thetab_i$ w/ prob  $1-\exp(-\alpha \tilde V(\thetab_i) \Delta t )$ }
  \Else {\ \textbf{if} \ $\tilde V(\thetab_i) < 0$ }  \State {
    duplicate $\thetab_i$ w/ prob  $1-\exp(-\alpha |\tilde V(\thetab_i)| \Delta t )$ } 
  \EndIf
  \EndFor
  \State $N_1$: total number of particles after the loop
  	\If {$N_1>N$} \State{kill $N_1 - N$ randomly selected particles}
  	\Else {\ \textbf{if} \ $N_1<N$ }  \State {  duplicate $N-N_1$ randomly selected particles}
	\EndIf
\EndWhile
\end{algorithmic}
\end{algorithm}

In the case of neural network parameter optimization, the birth-death
algorithm does not incur any significant computational cost beyond
regular stochastic gradient descent. Denoting the parameters
$\thetab_i=(c_i, \yb_i).$ and writing the neural network function as
\begin{equation}
\label{eq:nnwithc}
  f_n(\xb; c_1, \yb_1, \dots, c_n, \yb_n) = \frac1n \sum_{i=1}^n c_i
  \phi(\xb, \yb_i), 
\end{equation}
the potential $V(\thetab_i) = F(\thetab_i) + n^{-1} \sum_{j=1}^n 
  K(\thetab_i, \thetab_j)$ is given by
\begin{equation}
  V(\thetab_i) =  c_i \hat{V}(\yb_i)  \quad \text{with} \quad
  \hat{V}(\yb_i)  = \int_{\Omega} \phi(\xb,\yb_i)\left( 
    f_n(\xb; c_1, \yb_1, \dots, c_n, \yb_n)- f(\xb)\right) \nu(d\xb)
\end{equation}
Note that $\hat
V$ is the gradient of the loss with respect to the linear coefficient
vector $\partial_{c_i} V =
\hat{V}(\yb_i).$ 
Because we do not typically have access to the exact loss
function, the integrals required to compute
$\hat V$ are estimated using a finite number of data points.  Using a batch
of $P$ points in an update leads to an estimate $\hat V_P$ of
$\hat V$, which is used to determine the rate of killing/duplication.  In
this particular case, the only change to Algorithm~\ref{alg:scheme1}
is that the computation of $\tilde{V}$ is replaced with $c_i \hat
V_P(\yb_i) - n^{-1}\sum_{j=1}^nc_j \hat V_P(\yb_j)$ with
\begin{equation}
  \label{eq:empiricalpotential}
  \hat V_P(\yb_i) = \frac1P \sum_{p=1}^P \phi(\xb_p,\yb_i)\left( 
    f_n(\xb_p; c_1, \yb_1, \dots, c_n, \yb_n)- f(\xb_p)\right)  \qquad
  \{\xb_p\}_{p=1}^P = \text{batch}.
\end{equation}
Since this quantity is computed in the SGD update, the only additional
computation is the sum of $V_P$ over the
$n$ particles. 
The cost of the algorithm is $O(nP)$ at every
iteration.

For neural networks of the form given in Eq.~\eqref{eq:nnwithc} a
particularly simple modification of Algorithm~\ref{alg:scheme1}
enables particle creation from a prior distribution. The algorithm
proceeds through the initial birth-death loop as in
Algorithm~\ref{alg:scheme1}.  At the end of the initial loop, if the
total population has decreased, then additional particle are sampled
with configurations $(c,\yb)$ distributed according to the prior
distribution
\begin{equation}
  \label{eq:prior}
\mu_{\textrm b} (dc, d\yb) = \delta_0(dc) \bar \rho(\yb)d\yb
\end{equation}
so that a reinjected particle has zero contribution to the total
energy.

\medskip

\paragraph{Proximal Optimization: }
 Finally, let us note that it is possible to design algorithms for the
particles that mimic the proximal optimization scheme introduced in
\eqref{eq:proxi2}. For concreteness we focus on the cases of neural
networks---the ideas below can be easily adapted to the others
situations treated in this paper. Assume that the neural
representation at iterate $k$ is
\begin{equation}
  \label{eq:proxi8}
  f^{(n)}_k(\xb) = \frac1n \sum_{i=1}^n w_i^k \varphi(\xb,\thetab_i^k)
\end{equation}
where $\thetab_i^k$ denotes the parameter in the network and
$w_i^k\ge 0$ are extra weights satisfying $n^{-1}\sum_{i=1}^n w^k_i =1$---we
will define a dynamics for these weights in a moment. 
Notice that~\eqref{eq:proxi8} can be written as
\begin{equation}
  \label{eq:proxi8m}
  f^{(n)}_k (\xb) = \int_D \phi(\xb,\thetab) d\mu^{(n)}_k(\thetab), \qquad
  d\mu^{(n)}_k(\thetab) = \frac1n \sum_{i=1}^n w_i^k \delta_{\thetab_i^k}(d\thetab)
\end{equation}
and the loss is given by
\begin{equation}
  \label{eq:proxi10}
  \begin{aligned}
    \ell(\thetab_1^k,\ldots,\thetab_n^k;w_1^k,\ldots, w_n^k) &=
    \tfrac12 \EE_{y,\xb} |y - f^{(n)}_k(\xb) |^2 \\
    & = C_f + \frac1n\sum_{i=1}^n w_i^k F(\thetab_i^k) + \frac1{2n^2}
    \sum_{i,j=1}^n w_i^kw_j^k  K(\thetab_i^k,\thetab_j^k)
  \end{aligned}
\end{equation}
where $C_f = \tfrac12 \EE_{y} y^2$ and $F(\thetab)$ and
$K(\thetab,\thetab')$ given in~\eqref{fefe} and~\eqref{keke}, respectively.
The scheme we propose will update the $\thetab_i^k$ and the $w_i^k$
separately, the first by usual gradient descent over the loss, the
second by proximal gradient. That is, given $\{\thetab_i^k\}_{i=1}^n$
and $\{w_i^k\}_{i=1}^n$:

\smallskip

1. \textit{Gradient step.}  Evolve the parameters $\thetab_i^k$ by GD (or SGD if we
need to use the empirical loss) with the weights~$w_i^k$ kept
fixed. Do this for $m$ steps of size $\Delta t$ to obtain a new set of
$\{\thetab_i^{k+1}\}_{i=1}^n$.

\smallskip

2. \textit{Proximal step.} Evolve the weights $w_i^k$ with the
parameter $\thetab_i^{k+1}$ fixed using a proximal step based on the
particle equivalent of~\eqref{eq:proxi2}, i.e.
\begin{equation}
  \label{eq:proxi12}
  \{w_i^{k+1}\}_{i=1}^n \in \argmin \left(
    \ell(\thetab_1^{k+1},\ldots,\thetab_n^{k+1};w_1,\ldots, w_n)+
    \frac1{\tau n} \sum_{i=1}^n w_i \log(w_i/w_i^k) \right) 
\end{equation}
where the minimization is done under the constraint that $n^{-1} \sum_{i=1}^n
w_i = 1$. The equation for the minimizer $w_i^{k+1}$ is the discrete
equivalent of~\eqref{eq:proxi6}
\begin{equation}
  \label{eq:proxi13}
  w_i^{k+1} = C^{-1} w_i^k \exp\left( -\tau \tilde V_i^{k+1}\right) 
\end{equation}
where $C$ is a constant to be adjusted so that  $n^{-1}\sum_{i=1}^n
w^{k+1}_i = 1$ and
\begin{equation}
  \label{eq:proxi14}
  \tilde V_i^{k+1} = F(\thetab_i^{k+1} ) + \frac1n \sum_{j=1}^n
  w_j^{k+1} K(\thetab_i^{k+1},\thetab_j^{k+1}) 
\end{equation}
\eqref{eq:proxi13} is implicit in $w_i^{k+1}$ and should be solved by
iteration. Note that this proximal step is guaranteed to decrease the
loss. In practice, this step could eventually lead to big variations
of the weights. Should this happen, we add the additional step:

\smallskip

3. \textit{Resampling step.} Resample the weights
$\{w_i^{k+1}\}_{i=1}^n $ so as to keep them roughly equal to $1$
each, that is: eliminate the ones that are too small and transfer their
weights to the others: split the remaining (large) weights into bits
of size roughly $1$. There are standard ways to do this resampling
step that are unbiased and preserve the population size exactly. This
resampling step may increase the loss,  though not to leading
order. This step is the actual birth-death step in the scheme (and it
is also the only random component of it if the exact loss is used). 

\medskip

If we set $\tau= \alpha m \Delta t$ and set $\Delta t\to0$ and
$n\to\infty$, the scheme above is formally consistent with the PDE
\begin{equation}
  \label{eq:proxi15}
  \partial_t \mu_t = \nabla \cdot\left( \nabla V \mu_t \right) -
  \alpha V \mu_t + \alpha \bar V \mu_t .
\end{equation}
However, it is obviously not necessary to take either of these limits
explicitly in practice, and, as explained above, the proximal step is
guaranteed to decrease the loss. With a strict version of the the resampling 
step performed at
every iteration, in which the weights are taken to be in $\{0,1\}$ 
the scheme above recovers the one
described in Algorithm~\ref{alg:scheme1}. The main difference is that
in Algorithm~\ref{alg:scheme1} the proximal step~(\ref{eq:proxi13}) is
solved in one iteration, by substituting $w^{k+1}_i$ by $w_i^k$ at the
right hand side of~(\ref{eq:proxi13}).

Finally notice that if we were to implement the proximal step only and
skip both the gradient and the resampling steps, the scheme above is a
naive implementation of the lazy training scheme discussed
in~\cite{chizat:hal-01945578}. This highlights again why using
birth-death alone is not an efficient way to perform network
optimization, and it should be combined with standard GD.

 \section{Numerical Experiments}
\label{sec:experiments}

\subsection{Mixture of Gaussians}
We take as an illustrative example a mixture of Gaussians in dimension $d$,
\begin{equation}
  f(\xb) = \frac1m\sum_{i=1}^m \frac{\bar c_i}{(2\pi \sigma_i^2)^{d/2}} e^{ -|\xb-\bar \yb_i|^2/(2 \sigma_i^2)},
\end{equation}
which we approximate as a neural network with Gaussian nonlinearities with fixed standard deviation $\sigma<\min_i\sigma_i$,
\begin{equation}
  f_n(\xb; ; c_1, \yb_1, \dots, c_n, \yb_n) = \frac1n \sum_{i=1}^n
  \frac{c_i}{(2\pi \sigma^2)^{d/2}n} e^{-|\xb - \yb_i|^2/(2\sigma^2)},
\end{equation}
denoting the parameters $\thetab_i = (c_i, \yb_i).$
This is a useful test of our results because we can do exact gradient descent dynamics on the mean-squared loss function:
\begin{equation}
  \ell(c_1, \yb_1, \dots, c_n, \yb_n)
  = \frac12 \int_{\RR^d} \left|f(\xb)-
    f_n(\xb;  c_1, \yb_1, \dots, c_n, \yb_n)\right|^2 d\xb
\end{equation}
Because all the integrals are Gaussian, this loss can be computed analytically, and so can $\tilde V$ and its gradient.

In Fig. 1, we show convergence to the energy minimizer for a mixture of three Gaussians (details and source code are provided in the SM). 
The non-local mass transport dynamics dramatically accelerates convergence towards the minimizer. 
While gradient descent eventually converges in this setting---there is no metastability---the dynamics are particularly slow as the mass concentrates near the minimum and maxima of the target function.
However, with the birth-death dynamics, this mass readily appears at those locations.
The advantage of the birth-death dynamics with a reinjection distribution $\mu_{\textrm{b}}$ is highlighted by choosing an unfavorable initialization in which the particle mass is concentrated around $y=-2.$
In this case, both GD and GD with birth-death~\eqref{eq:pde_bd} do not converge on the timescale of the dynamics. 
With the reinjection distribution, new mass is created near $y=2$ and convergence is achieved.

\begin{figure}
\begin{center}
  \includegraphics[width=\linewidth]{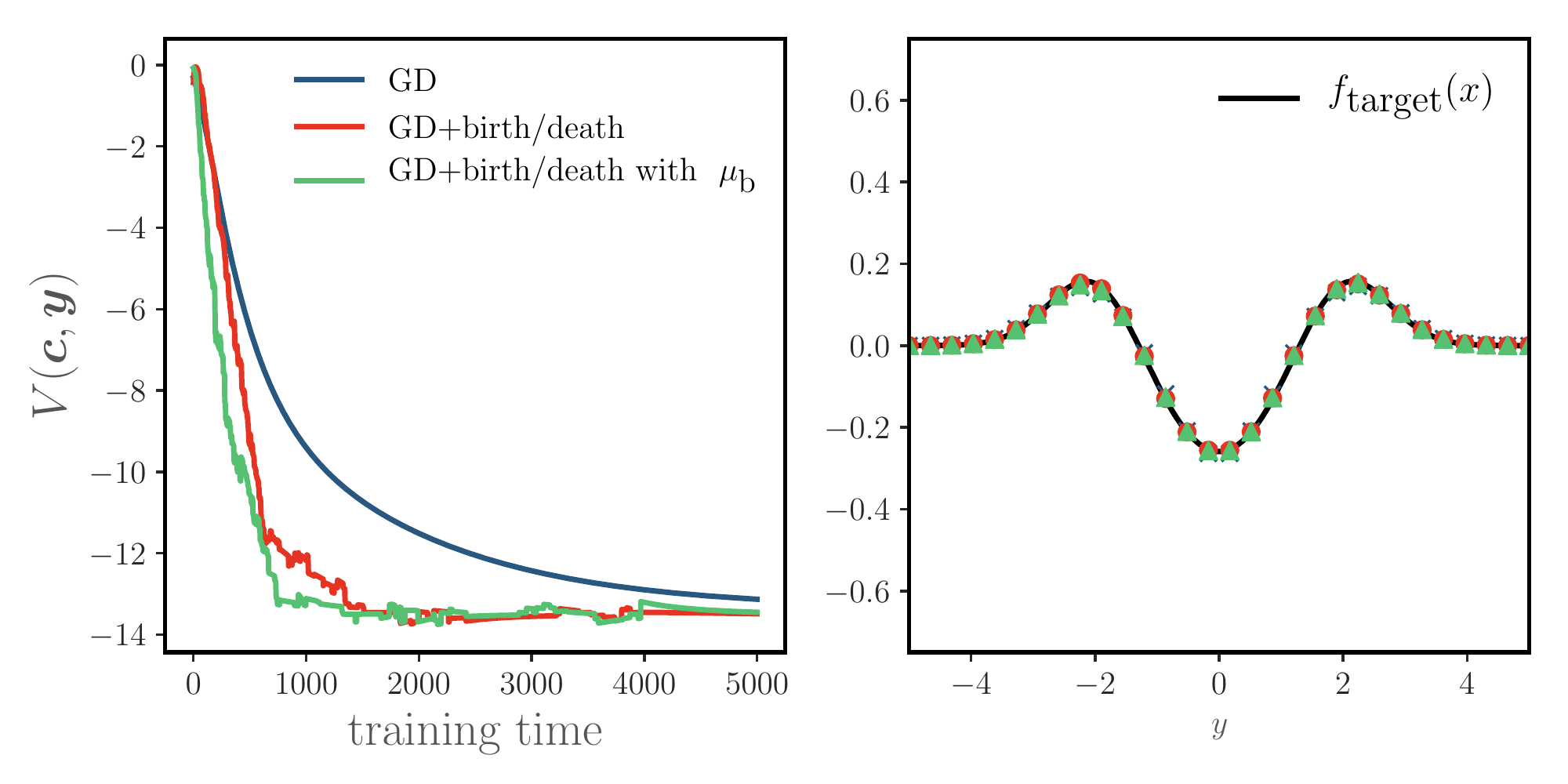}
  \includegraphics[width=\linewidth]{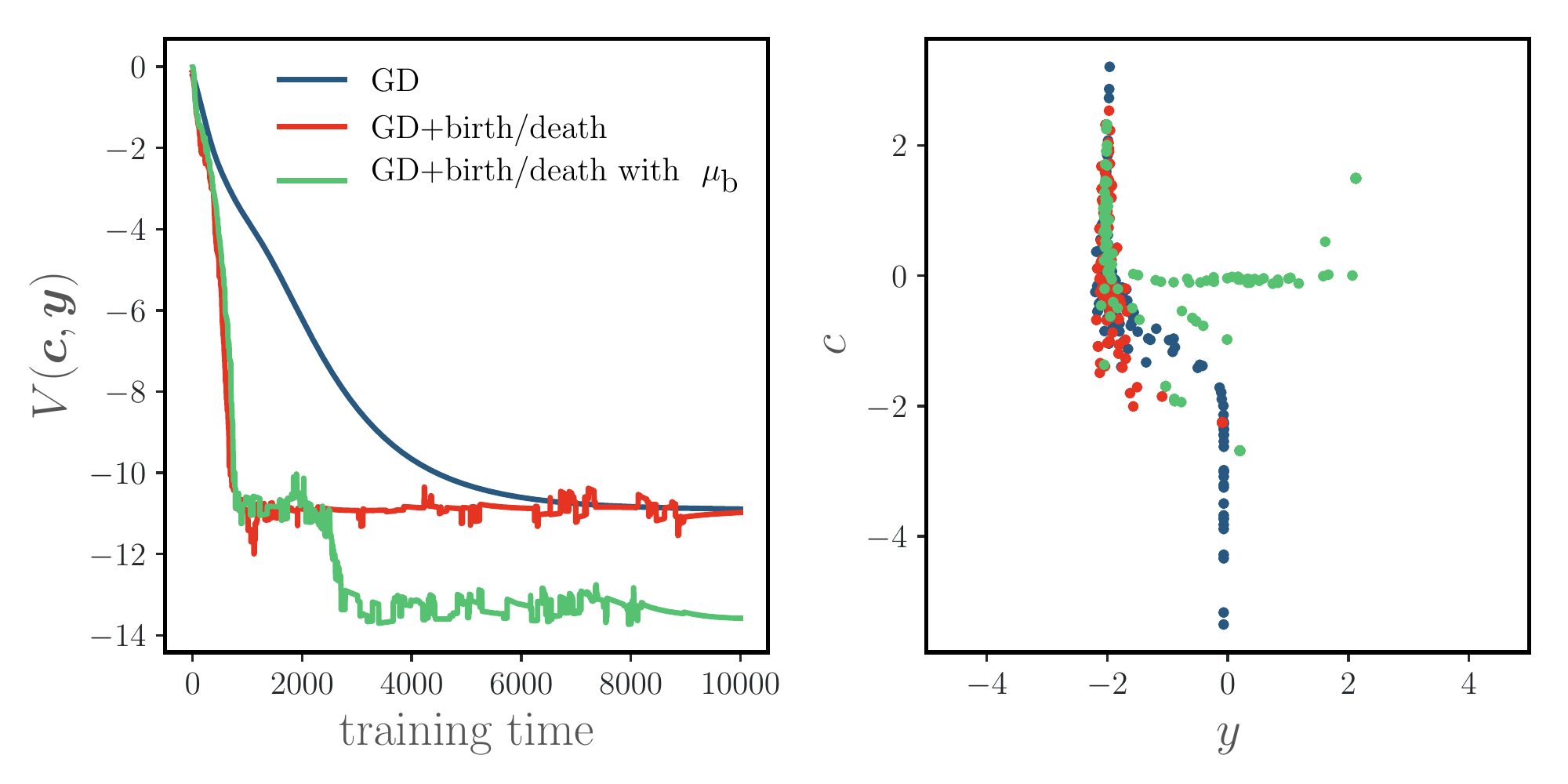}
 \end{center}
  \caption{Top left: Convergence of the gradient descent dynamics without birth-death, with birth-death, and using a reinjection distribution. Top right: For appropriate initialization, the three dynamical schemes all converge to the target function. Bottom left: For bad initialization (narrow Gaussian distributed around y=-2), GD and GD+birth-death do not converge on this timescale. Interestingly, with the reinjection via distribution $\mu_\textrm{b}$, convergence to the global minimum is rapidly achieved. Bottom right: The configuration of the particles in $\thetab=(y,c)$. Only with the reinjection distribution does mass exist near $y=2$.}
\end{figure}

\subsection{Student-Teacher ReLU Network}\label{sec:stnet}

In many optimization problems, it is not possible to evaluate $\tilde V$ exactly.
Instead, typically $\tilde V$ is estimated as a sample mean over a batch of data.
We consider a student-teacher set-up similar to \cite{chizat:hal-01945578} in which we use single hidden layer ReLU networks to approximate a network of the same type with fewer neurons.
We use as the target function a ReLU network with 50-$d$ input and 10 hidden units.
We approximate the teacher with neural networks with $n=50$ neurons (see SM). 
The networks are trained with stochastic gradient descent (SGD) and the mini-batch estimate of the gradient of output layer, which is computed at each step of SGD, is used to compute $\tilde V,$ which determines the rate of birth-death.
In experiments with the reinjection distribution, we use \eqref{eq:prior} with Gaussian $\bar \rho.$

\begin{figure}
  \begin{center}
  \includegraphics[width=0.6\linewidth]{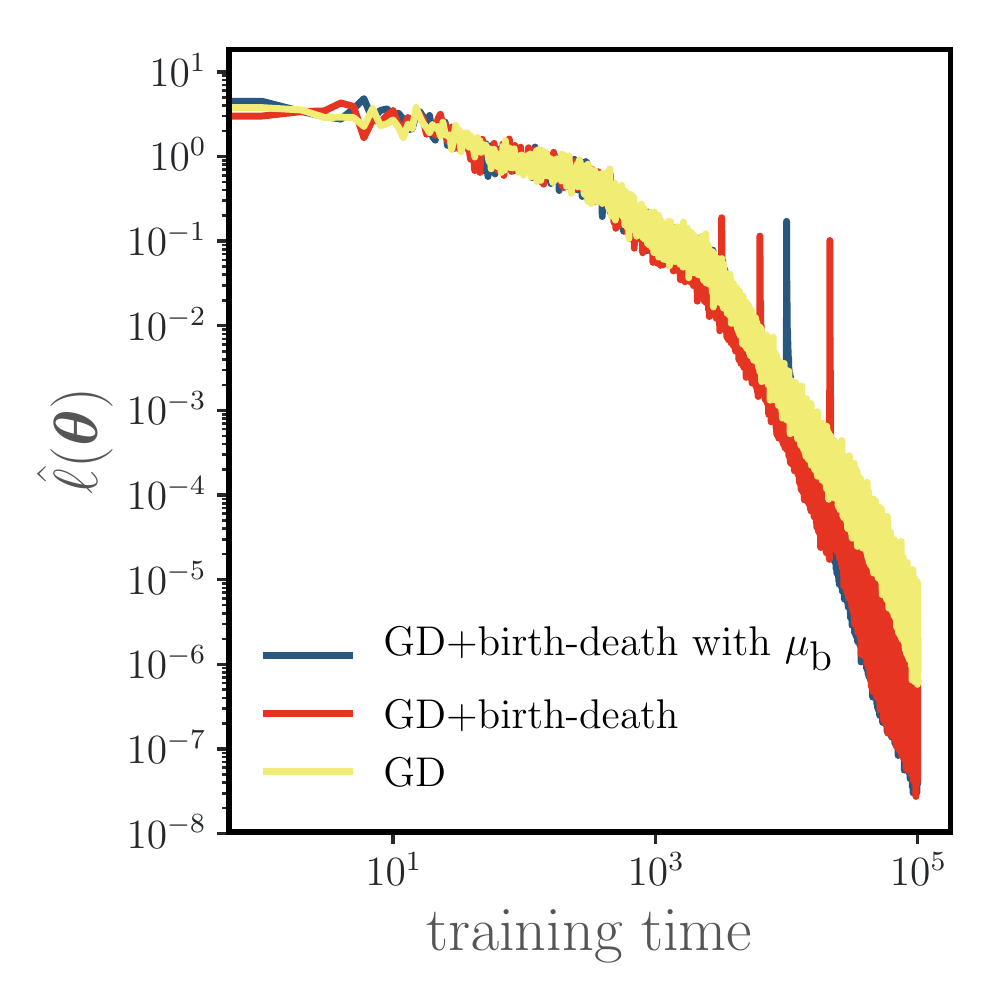}
  \end{center}
  \caption{The batch loss as a function of training time for the student-teacher ReLU network described in Sec.~\ref{sec:stnet}. The birth-death dynamics accelerates convergence, both with and without the reinjection distribution.}
  \label{fig:nn}
\end{figure}

As shown in Fig.~\ref{fig:nn}, we find that the birth-death dynamics accelerates convergence to the teacher network. 
We emphasize that because the birth-death dynamics is stochastic at finite particle numbers, the fluctuations associated with the process could be unfavorable in some cases.
In such situations, it is useful to reduce $\alpha$ as a function of time. 
On the other hand, in some cases we have observed much more dramatic accelerations from the birth-death dynamics. 
 
\section{Conclusions}

The success of an optimization algorithm based on gradient descent requires good coverage of the parameter space so that local updates can reach the minima of the loss function quickly.
Our approach liberates the parameters from a purely local dynamics and allows rapid reallocation to values at which they can best reduce the approximation error. 
Importantly, we have constructed the non-local birth-death dynamics so that it converges to the minimizers of the loss function.
For a very general class of minimization problems---both interacting and non-interacting potentials---we have established convergence to energy minimizers under the dynamics described by the mean-field PDE with birth-death.
Remarkably, for interacting systems with we can guarantee global convergence for sufficiently regular initial conditions. 
We have also computed the asymptotic rate of convergence with birth-death dynamics. 

These theoretical results translate to dramatic reductions in convergence time for our illustrative examples. 
It is worth emphasizing that the schemes we have described are straightforward to implement and come with little computational overhead.
Extending this type of dynamics to deep neural network architectures could accelerate the slow dynamics at the initial layers often observed in practice. 
Hyperparameter selection strategies based on evolutionary algorithms~\cite{such2017deep} provide another interesting potential application of our approach. 

While we have characterized the basic behavior of optimization under the birth-death dynamics, many theoretical questions remain. First, we did not address generalization; understanding the role of the extra birth/death term in controlling the generalization gap is an important future question, in particular relating it to the lazy-training regime of \cite{chizat:hal-01945578}. Next, we need to assume the existence of weak solutions through (\ref{eq:15}) with an initial measure $\mu_0$ that has full support, yet it may be possible to certify that the dynamics exist for all times if $\mu_0$ decays sufficiently fast. 
Besides, more explicit calculations of global convergence rates for the interacting case and tighter rates for the non-interacting case would be exciting additions. 
The proper choice of $\mu_{\textrm{b}}$ is another question worth exploring because, as highlighted in our simple example, favorable reinjection distributions can rapidly overcome slow dynamics. 
Finally, a mean-field perspective on deep neural networks would enable us to translate some of the guarantees here to deep architectures.

\section*{Acknowledgments}
We would like to acknowledge the useful and detailed comments by Sylvia Serfaty and Yann Ollivier on previous versions of this manuscript.

\bibliography{refs2}
\bibliographystyle{alpha}

\newpage
\appendix

\section{Generalizations of~(\ref{eq:pde_control})}
\label{sec:modify}

Here we mention two ways in which we can
modify~(\ref{eq:pde_control}) to certain advantages. For example, we
can replace this equation with
\begin{equation}
  \partial_t \mu_t = \div \left( \mu_t \grad V \right) - \alpha
  f(V-\bar V) \mu_t - \bar f \mu_t,
  \label{eq:pde_control2}
\end{equation}
where $f: \RR \to \RR$ is some function and
$\bar f = \int_D f(V-\bar V) d\mu_t$. As we will see in
Proposition~\ref{prop:consis2}, as long as $zf(z)\ge0$ for all
$z \in \RR$, the additional term in~(\ref{eq:pde_control2}) increase
the rate of decay of the energy. 

While the birth-death dynamics described above ensures convergence in the mean-field limit, when $n$ is finite, particles can only be created in proportion to the empirical distribution $\mu^{(n)}.$
In particular, such a birth process corresponds to ``cloning'' or creating identical replicas of existing particles.
In practice, there may be an advantage to exploring parameter space with a distribution distinct from the instantaneous empirical particle distribution~\eqref{eq:rhon}.
To enable this exploration we introduce a birth term proportional to a distribution $\mu_{\textrm{b}}$ which we will assume has full support on $D$.
In this case, the time evolution of the distribution is described by
\begin{equation}
\label{eq:pde_prior}
\begin{aligned}
  \partial_t \mu_t = \div \left( \mu_t \grad V \right) & - \alpha
  (V-\bar V)_+ \mu_t + \alpha \left( {\textstyle \int_D (V-\bar
      V)_+d\mu_t}\right)
  \frac{\mu_{\textrm{b}}\mathbbm{1}_{V\le \bar V }}{\mu_b(V \le \bar V)} \\
  & + \alpha' (V-\bar V)_- \mu_{\textrm{b}} - \alpha' \left(
    {\textstyle \int_D (V-\bar V)_-d\mu_{\text{b}}}\right) \frac{\mu_t
    \mathbbm{1}_{V> \bar V}}{\mu_t(V> \bar V)},
\end{aligned}
\end{equation}
where $\alpha,\alpha'>0$, $(V-\bar V)_+=\max(V-\bar V,0)\ge0$,
$(V-\bar V)_-=\max(\bar V-V,0)\ge0$.  That is, we kill particles in
proportion to $\mu_t$ in region where $V>\bar V$ but create new
particles from $\mu_{\textrm{b}}$ in regions where $V\le\bar V$. We
could also combine~(\ref{eq:pde_control2}) with (\ref{eq:pde_prior})
to obtain other variants.

These alternative birth-death dynamical schemes also satisfy the consistency conditions of Proposition \ref{prop:consis1}:
\begin{proposition}
\label{prop:consis2}
Let $\mu_t$ be a solution of~(\ref{eq:pde_control2}) with $f$ such
that $zf(z)\ge 0$ for all $z\in \RR$ or (\ref{eq:pde_prior}), with
$\mu_0 \in \mathcal{M}(D)$.  Then, $\mu_t(D)=1$ for all $t\ge0$, and
${E}(t)=\mathcal{E}[\mu_t]$ satisfies
\begin{equation}
\label{eq:fifi}
    \dot{{E}}(t) \leq - \int_D | \nabla V(\thetab,[\mu_t])|^2 \mu_t(d\thetab) \,.
\end{equation}
\end{proposition}
{\it Proof:} 
By considering again $1$ and $V(\cdot,[\mu_t])$ as a test function
in~(\ref{eq:pde_control2}) or (\ref{eq:pde_prior}), we verify that
$\partial_t\mu_t(D) =0$. In addition, (\ref{eq:pde_control2}) implies
that
\begin{equation*}
\begin{aligned}
  \dot{E}(t) &=  \int_D V(\thetab,[\mu_t]) \partial_t\mu_t(d\thetab) \\
  &=  \int_D \left( V(\thetab,[\mu_t])- \bar V[\mu_t]\right) \partial_t\mu_t(d\thetab) \\
&= - \int_D | \nabla V |^2 d\mu_t 
 - \alpha \int_D (V-\bar V) f(V-\bar V) d\mu_t 
\end{aligned}
\end{equation*}
which proves (\ref{eq:fifi}) for (\ref{eq:pde_control2}) since all the
terms at the right hand side of this equation are negative
individually if $zf(z) \ge 0 $ for all $z\in \RR$.
Similarly,  (\ref{eq:pde_prior}) implies that
\begin{equation*}
\begin{aligned}
  \dot{E}(t) &=  \int_D V(\thetab,[\mu_t]) \partial_t\mu_t(d\thetab) \\
  &=  \int_D \left( V(\thetab,[\mu_t]) -\bar V[\mu_t]\right) \partial_t\mu_t(d\thetab) \\
  &= - \int_D | \nabla V |^2 d\mu_t - \alpha \int_D (V -\bar V)^2_+
  d\mu_t - \alpha \frac{\int_D (V-\bar V)_+ d\mu_t \int_D (V-\bar V)_-
    d\mu_{\textrm{b}}}
  {\mu_b(V \le 0)} \\
  &\quad -\alpha' \int_D (V-\bar V)^2_{-} d\mu_{\textrm{b}}
  -\alpha'\frac{\int_D (V-\bar V)_{-} d\mu_{\textrm{b}} \int_D (V-\bar
    V)_+d\mu_t}{\mu_t(V> 0)},
\end{aligned}
\end{equation*}
which proves (\ref{eq:fifi}) for (\ref{eq:pde_prior}) since all the
terms at the right hand side of this equation are negative.\hfill
$\square$

\section{Convergence and  Rates in the Non-interacting Case}
\label{sec:convergence-noninteracting}

\subsection{Non-interacting Case without the Transportation Term}
\label{subsec:non_interacting_BD}

Let us look first at the PDE satisfied by the measure $\mu$ in the
non-interacting case, i.e. with $V = F$ satisfying
Assumption~\ref{as:noninter}, and without the transportation term:
\begin{equation}
  \partial_t \mu_t = - \alpha F(\thetab ) \mu_t + \alpha \bar{F}(t) \mu_t,
\end{equation}
where $\bar F(t) = \int_{\RR^k} F(\thetab) \mu_t(d\thetab)$.
This equation can be solved exactly. Assuming that 
$\mu_0$ has a density everywhere positive on $\RR^k$, $\mu_t$
has a density $\rho_t$ given by
\begin{equation}\label{eq:mu_tmp_BD}
  \rho_t(\thetab ) =e^{\alpha \int_0^t\bar{F}(s)ds -\alpha t F(\thetab )} \rho_0(\thetab ).
\end{equation}
The normalization condition $\mu_t(\RR^k) =\int_{\RR^k}
\rho_t(\thetab) d\thetab= 1$ leads to: 
\begin{align*}
   &e^{\alpha \int_0^t\bar{F}(s)ds } \int_{\RR^k} e^{-\alpha t
     F(\bm{\theta'})}
     \rho_0(\thetab')d\thetab'=1\\
  \Leftrightarrow \quad & e^{-\alpha \int_0^t\bar{F}(s)ds } =
                   \int_{\RR^k} e^{-\alpha t F(\bm{\theta'})} \rho_0(\thetab')d\thetab'.
\end{align*}
Therefore, by plugging this last expression in equation~\eqref{eq:mu_tmp_BD}, we obtain the explicit expression
\begin{equation}
\rho_t(\thetab ) =\frac{e^{-\alpha t F(\thetab )} \rho_0(\thetab )}{\int_{\RR^k} e^{-\alpha t F(\thetab ')} \rho_0(\thetab')d\thetab '}.
\end{equation}
We can use this equation to express the energy
$\bar F(t) = \int_{\RR^k} F(\thetab) \rho_t(\thetab)d\thetab$:
\begin{equation}
  \begin{aligned}
    \bar F(t)= \frac{\int_{\RR^k} F(\thetab )e^{-\alpha t F(\thetab )}
      \rho_0(\thetab)d\thetab}{\int_{\RR^k} e^{-\alpha t F(\thetab)}
      \rho_0(\thetab)d\thetab} = \frac{d}{d\alpha t}G(\alpha t),
  \end{aligned}
\end{equation}
where $G(\alpha t)$ is the function defined as: 
\begin{equation}
G(\alpha t) = - \log \int_{\RR^k} e^{-\alpha t F(\thetab )}  \rho_0(\thetab)d\thetab .    
\end{equation}

At late times, the factor $e^{-\alpha t F(\thetab )}$ focuses all the
mass in the vicinity of the global minimum of $F$. Therefore, we can
neglect the influence of the density $\rho_0$ in this integral. More
precisely a calculation using the Laplace method indicates that
\begin{equation}
  \begin{aligned}
    & \int_{\RR^k} e^{-\alpha t F(\thetab )} d\thetab   \sim
    (2\pi)^{d/2}(\alpha t)^{-d/2}(\det(H^*))^{-1/2}.
  \end{aligned}
\end{equation}
where $H^*= \nabla\nabla F(\thetab ^*)$ is the Hessian at the global
minimum located at~$\thetab^*$, and $\sim $ indicates that the ratio
of both sides of the equation tend to 1 as $\alpha t \to\infty$.  This shows that
\begin{equation}
    \bar F(t) \sim \tfrac{1}{2}d(\alpha t)^{-1} \qquad \text{as \ \ $\alpha t\to \infty$}
\end{equation}

\subsection{Non-interacting Case with Transportation and Birth-death}

\subsubsection{Proof of Theorem~\ref{th:local-non-rate}}\label{subsec:proof_local_rate_thm}

We first prove the following intermediate result
\begin{lemma}
\label{le1}
Let $\delta>0$ arbitrary, and define
$$\phi_\delta(\thetab)= \max(0, 1- \delta^{-1} F(\thetab))~,~f_\delta=\int_{\RR^k} \phi_\delta(\thetab) \mu_0(d\thetab)~.$$
Then
\begin{equation}
\label{lu0}
 \forall t \ : \quad   E(t) \leq \delta + \frac{1}{\alpha t f_\delta}~.
\end{equation}
\end{lemma}
{\it Proof:}
By slightly abusing notation, we define 
$$f_\delta(t)=\int_{\RR^k}  \phi_\delta(\thetab) \mu_t(d\thetab)~.$$
We consider the following Lyapunov function:
\begin{equation}
\label{lu1}
    \mathcal{L}_\delta(t) = \alpha t (E(t) - \delta) + \frac{1}{f_\delta(t)}~.
\end{equation}
Its time derivative is 
\begin{equation}
\label{lu2}
\dot{\mathcal{L}}_\delta(t) = \alpha(E(t) - \delta) + \alpha t \dot{E}(t) - \frac{\dot{f_\delta}(t)}{f_\delta^2(t)} ~.
\end{equation}
By definition, we have 
\begin{equation}
\label{lu3}
    \dot{E}(t) = -\int_{\RR^k}  | \nabla F(\thetab)|^2 \mu_t(d\thetab) - \alpha \int_{\RR^k}  (F(\thetab) - F(t))^2 \mu_t(d\thetab) \leq 0~.
\end{equation}
We also have 
\begin{align}
\label{lu4sm}
\dot{f}_\delta(t) &= -\int_{\RR^k}  \langle \nabla \phi_\delta(\thetab), \nabla F(\thetab) \rangle \mu_t(d\thetab) - \alpha \int_{\RR^k}  \phi_\delta(\thetab) F(\thetab)\mu_t(d\theta) + \alpha E(t) f_\delta(t) \nonumber \\ 
&\geq \delta^{-1} \int_{\RR^k} | \nabla F(\thetab)|^2 \mu_t(d\thetab) + \alpha (E(t) - \delta) f_\delta(t) \nonumber \\
&\geq \alpha (E(t) - \delta) f_\delta(t)~.
\end{align}
Observe that $0 \leq f_\delta(t) < 1$ because otherwise $F$ would be flat (in which case the energy is $0$). 
Also, we can assume wlog that $E(t)-\delta > 0$, since otherwise the statement of the lemma is trivially verified.
By plugging (\ref{lu3}) and (\ref{lu4sm}) into (\ref{lu2}) we have 
\begin{equation}
 \dot{\mathcal{L}}_\delta(t) \leq  \alpha(E(t) - \delta) - \alpha (E(t)-\delta) f_\delta^{-1}(t) = \alpha (E(t) - \delta)(1 - f_\delta^{-1}(t)) \leq 0~.
\end{equation}
Finally, since $f^{-1}_\delta(t) \geq 0$, we have 
$$(E(t) - \delta) \leq \frac{\mathcal{L}_\delta(t)}{\alpha t} \leq \frac{\mathcal{L}_\delta(0)}{\alpha t} 
= \frac{1}{\alpha t f_\delta}~,$$
which concludes the proof of the Lemma. \hfill$\square$ 

{\it Proof of Theorem~\ref{th:local-non-rate}:}
In order to prove (\ref{lu00}), 
we apply the previous lemma for $\delta \to 0$.
Let $\theta^* = \arg\min V(\theta)$, 
We have 
$F(\thetab^*)  = 0$, and $\| \nabla\nabla F(\thetab) \| \leq \beta$ for some $\beta >0$.
Then, for $\delta$ sufficiently small, 
the indicator function $\phi_\delta(\thetab)$ is localized in the set 
$$\left\{ \thetab \in \mathbb{R}^k; \tfrac{1}{2}\<(\thetab-\thetab^*), H^* (\thetab-\thetab^*) \leq \delta \right\} \supseteq \left\{ \thetab \in \mathbb{R}^d; \| \thetab - \thetab^*\|^2 \leq 2\beta^{-1}\delta\right\}~.$$
where $H^*=\nabla \nabla F(\thetab^*)$.
It follows that for sufficiently small $\delta$,
\begin{align}
\label{lu4}
f_\delta &= \int_{\RR^k}  \phi_\delta(\thetab) \mu_0(d\thetab) \nonumber \\
&\gtrsim \rho_0(\thetab^*) \int_{\| \thetab - \thetab^*\| \leq \sqrt{2\beta^{-1}\delta}} \left(1 - \tfrac{1}{2}\delta^{-1}\<(\thetab-\thetab^*), H^* (\thetab-\thetab^*) \>\right) d\thetab \nonumber \\
&\sim \rho_0(\thetab^*) \left(2\beta^{-1}\delta\right)^{d/2}~.
\end{align}
By plugging (\ref{lu4}) into (\ref{lu0}) we obtain 
$$\forall\,\delta,t>0 \ : \qquad ~E(t) \leq \delta + \frac{1}{\alpha t} \left(\frac{\beta}{2\delta}\right)^{d/2} \sim \delta + C\delta^{-d/2}t^{-1}~,$$
which implies that in order to reach an error $\epsilon$, we need 
\begin{equation*}
t_\epsilon = O\left( \epsilon^{-(d+2)/2}\right)~,    
\end{equation*}
which shows~\eqref{lu00}.

To obtain the asymptotic convergence rate in~\eqref{eq:expdecay}, note
that by Lemma~\ref{lem:char} below the energy
$\bar F(t) = \int_{\RR^k} F(\thetab) \rho_t(\thetab) d\thetab$ can be
written in terms of~\eqref{eq:solPDE} as
\begin{equation}
  \label{eq:barF-interact}
  \bar F(t) = \frac{\int_{\RR^k} F(\thetab) \exp\left(\int_{-t}^0 (-\alpha F(\Thetab(s,\thetab ))
      +\Delta F(\Thetab(s,\thetab )))ds \right)  \rho_0(\Thetab(-t,\thetab ))d\thetab}{
    \int_{\RR^k} \exp\left(\int_{-t}^0 (-\alpha F(\Thetab(s,\thetab ))
      +\Delta F(\Thetab(s,\thetab )))ds \right)  \rho_0(\Thetab(-t,\thetab ))d\thetab}
\end{equation}
For large $t$, we can again use Laplace method to confirm that
$\rho(t,\thetab )$ concentrates near the absolute minimum of
$F(\thetab) $ located at $\thetab^*$. To see why notice that
$\Thetab(t,\thetab)$ converge, as $t\to\infty$, near local minima of
$F$. Suppose that these minima are located at $\thetab_1^*=\thetab^*$,
$\thetab_2^*$, etc. At these minima we have $\nabla F(\thetab^*_j)=0$,
and if in~(\ref{eq:charact}) we replace $F(\thetab)$ by its quadratic
approximation around any $\thetab_j^*$,
$\tfrac12 \< \thetab -\thetab_j ^*, H_j^*(\thetab -\thetab_j ^*)\>$
with $ H_j^* = \nabla \nabla H(\thetab_j^*)$ positive definite, the
solution to this equation reads
\begin{equation}
  \Thetab^j_{\text{quad}} (t,\thetab )=\thetab_j ^* + e^{-H^*t}(\thetab -\thetab_j ^*) 
\end{equation}
from which we deduce
\begin{equation}
  \int_{-t}^0\Delta F(\Thetab^j_{\text{quad}}  (s,\thetab )) ds =  \tr(H_j^*) t,
\end{equation}
and 
\begin{equation}
    \begin{aligned}
      -\alpha \int_{-t}^0 F(\Thetab^j_{\text{quad}} (s,\thetab ))ds &=
      \alpha F(\thetab^*_j) t
      -\tfrac{1}{2}\alpha \int_{-t}^0
      \langle \tilde{\thetab}_j,e^{-H^*s}H^*e^{-H^*s}\tilde{\thetab}_j\rangle\\
      &= \alpha F(\thetab^*_j) t - \tfrac{1}{4}\alpha \langle
      \tilde{\thetab}_j,(e^{2H^*t}-\mathrm{Id}) \tilde{\thetab}_j\rangle.
    \end{aligned}
\end{equation}
where $\tilde{\thetab}_j = \thetab -\thetab_j ^*$. Since
$F(\thetab_j^*)>0$ except for the the global minimum
$F(\thetab_1^*)=F(\thetab_1^*)=0$, for large $t$, the only points that
contribute to the integrals in~\eqref{eq:barF-interact} are those in
a small region near $\thetab^*$ where we can replace
$\Thetab(t,\thetab)$ by $\Thetab^1_{\text{quad}}(t,\thetab)$. As a
result we can again neglect $\rho_0$ in these integrals, and evaluate
them as if $\rho_t$ was asymptotically the Gaussian density:
\begin{equation}
    \rho_t(\thetab ) \sim \mathcal{N}(\thetab ^*,2\alpha^{-1}e^{-2H^*t}).
\end{equation}
This quantifies the late stages of the global convergence to the
minimum and confirms the asymptotic decay rate in~\eqref{eq:expdecay},
thereby concluding the proof of
Theorem~\ref{th:local-non-rate}. \hfill $\square$

\begin{lemma}
\label{lem:char}
Denote by $\Thetab(t,\thetab)$ the solution of the ODE
\begin{equation}
  \dot{\Thetab}(t,\thetab) = -\grad F(\Thetab(t,\thetab)), \qquad
  \Thetab(0,\thetab) = \thetab
\end{equation}
Then under the conditions of Theorem~\ref{th:local-non-rate}, the
solution $\mu_t$ of the PDE~\eqref{eq:pde_bd}  has a density
$\rho_t$ given by
\begin{equation}
\label{eq:solPDE}
\begin{aligned}
  \rho_t(\thetab) = \frac{\exp\left(\int_{-t}^0 G(\Theta(s,\thetab)) ds \right)
    \rho_0(\Thetab(-t,\thetab))} {\int_D \exp\left(\int_{-t}^0
      G(\Theta(s,\thetab')) ds\right)\rho_0(\Thetab(-t,\thetab'))d\thetab'}
\end{aligned}
\end{equation}
where $G(\thetab) = \Delta F(\thetab)-\alpha F(\thetab)$.
\end{lemma}

\textit{Proof:}
Since the initial $\mu_0$ has a density $\rho_0>0$, so does
$\mu_t$ for all $t>0$ (but not in the limit as $t\to\infty$) and its density
satisfies
\begin{equation}
\label{eq:pde_control_noninteracting} 
\begin{aligned}
    \partial_t \rho_t &= \div \left( \rho_t \grad F(\thetab )\right) - \alpha F(\thetab ) \rho_t + \alpha \bar F(t)  \rho(t),
\end{aligned}
\end{equation} If
$\Thetab(t,\thetab )$ satisfies
\begin{equation}
  \label{eq:charact}
  \dot{\Thetab}(t,\thetab )= -\nabla F(\Thetab(t,\thetab ))\qquad
  \Thetab(0,\thetab )=\thetab .
\end{equation}
we have
\begin{equation}
  \begin{aligned}
    \frac{d}{dt} \rho_t(\Thetab(t,\thetab ))
    & = \partial_t \rho_t(\Thetab(t,\thetab )) +
    \dot{\Thetab}(t,\thetab )) \cdot \grad \rho_t(\Thetab(t,\thetab ))\\
    &= \Delta F(\Thetab(t,\thetab )) \rho(t,\Thetab(t,\thetab ))-
    \left(F(\Thetab(t,\thetab )) - \alpha \bar F(t)\right)
    \rho_t(\Thetab(t,\thetab )).
  \end{aligned}
\end{equation}
Therefore
\begin{equation}
  \rho_t(\Thetab(t,\thetab ))=\exp\left(\int_0^t (-\alpha
    F(\Thetab(s,\thetab ))+\alpha \bar F(s)
  +\Delta F(\Thetab(s,\thetab )))ds \right)  \rho_0(\thetab ).    
\end{equation}
By using $\Thetab(t,\Thetab(s,\thetab)) = \Thetab(t+s,\thetab)$ and
the normalization condition, this implies
\begin{equation}
\label{eq:rhoexplicit}
  \rho_t(\thetab)=\frac{\exp\left(\int_{-t}^0 (-\alpha F(\Thetab(s,\thetab ))
  +\Delta F(\Thetab(s,\thetab )))ds \right)  \rho_0(\Thetab(-t,\thetab ))}{
  \int_{\RR^k} \exp\left(\int_{-t}^0 (-\alpha F(\Thetab(s,\thetab '))
  +\Delta F(\Thetab(s,\thetab ')))ds \right)  \rho_0(\Thetab(-t,\thetab '))d\thetab'}.    
\end{equation}
This is~\eqref{eq:solPDE} and terminates the proof of the lemma.\hfill
$\square$

\section{Derivation of \eqref{eq:euler_lagrange}}
\label{sec:euler-lagrange}

Let $\mu_*$ be a minimizer and compare its energy to that of any other
probability measure $\mu$. Since the energy minimum is unique by
convexity, we must have $\mathcal{E}[\mu] \ge \mathcal{E}[\mu_*]$.  A
direct calculation shows that
\begin{equation}
  \label{eq:1el}
  \begin{aligned}
    \mathcal{E}[\mu] &= \mathcal {E}[\mu_*] +
     \int_D V(\thetab,[\mu_*]) (\mu(d\thetab) - \mu_*(d\thetab))
    \\
    & \quad+ \tfrac12 \int_{D\times D} K(\thetab,\thetab')
    (\mu(d\thetab) - \mu_*(d\thetab)) (\mu(d\thetab') -
    \mu_*(d\thetab'))
  \end{aligned}
\end{equation}
The last term at the right hand side is always non-negative. Focusing
on the second term, if we denote $\supp \mu_* = D_*$, we can write it
as
\begin{equation}
  \label{eq:1el2}
  \begin{aligned}
    \int_D V(\thetab,[\mu_*]) (\mu(d\thetab) - \mu_*(d\thetab))
    & = \int_D \left( V(\thetab,[\mu_*]) - \bar V [\mu_*])
    \right) (\mu(d\thetab) - \mu_*(d\thetab))\\
    & = \int_{D_*^c} \left( V(\thetab,[\mu_*]) - \bar V [\mu_*])
    \right) \mu(d\thetab)
  \end{aligned}
\end{equation}
where we used $V(\thetab,[\mu_*]) = \bar V [\mu_*]$ on $D_*$ and
$\mu_*=0 $ on $D_*^c$. The only possibility to make this term nonnegative
for all $\mu$ is to have $V(\thetab,[\mu_*]) \ge \bar V [\mu_*]$ on
$D_*^c$.

\section{Proof of Theorem~\ref{th:global}}
\label{app:proof_global_convergence}

We begin by noting that, if \eqref{eq:15} holds for al $t>0$, then $\bar V[\mu_t] = -\alpha^{-1} d\log C(t)/dt$ must be well-defined at all times. From~\eqref{eq:18}, this derivative is given by
\begin{equation}
    \bar V[\mu_t] = -\alpha^{-1}\frac{d}{dt} \log C(t) =  \frac{\int_D V(\Thetab(t,\thetab),[\mu_t]) e^{-\alpha \int_0^t V(\Thetab(s,\thetab),[\mu_s]) ds} \mu_0(d\thetab)}{\int_D e^{-\alpha \int_0^t V(\Thetab(s,\thetab),[\mu_s]) ds} \mu_0(d\thetab)}
\end{equation}
Differentiating one more times gives
\begin{equation}
    \begin{aligned}
    \frac{d}{dt} \bar V[\mu_t]  &= -\alpha \frac{\int_D |V(\Thetab(t,\thetab),[\mu_t])|^2 e^{-\alpha \int_0^t V(\Thetab(s,\thetab),[\mu_s]) ds} \mu_0(d\thetab)}{\int_D e^{-\alpha \int_0^t V(\Thetab(s,\thetab),[\mu_s]) ds} \mu_0(d\thetab)}\\
    &\quad +\alpha \left(\frac{\int_D V(\Thetab(t,\thetab),[\mu_t]) e^{-\alpha \int_0^t V(\Thetab(s,\thetab),[\mu_s]) ds} \mu_0(d\thetab)}{\int_D e^{-\alpha \int_0^t V(\Thetab(s,\thetab),[\mu_s]) ds}}\right)^2\\
    &\quad + \frac{\int_D \partial_t V(\Thetab(t,\thetab),[\mu_t]) e^{-\alpha \int_0^t V(\Thetab(s,\thetab),[\mu_s]) ds} \mu_0(d\thetab)}{\int_D e^{-\alpha \int_0^t V(\Thetab(s,\thetab),[\mu_s]) ds} \mu_0(d\thetab)} \\
    &= -\alpha \frac{\int_D |V(\Thetab(t,\thetab),[\mu_t])|^2 e^{-\alpha \int_0^t V(\Thetab(s,\thetab),[\mu_s]) ds} \mu_0(d\thetab)}{\int_D e^{-\alpha \int_0^t V(\Thetab(s,\thetab),[\mu_s]) ds} \mu_0(d\thetab)}\\
    &\quad +\alpha \left(\frac{\int_D V(\Thetab(t,\thetab),[\mu_t]) e^{-\alpha \int_0^t V(\Thetab(s,\thetab),[\mu_s]) ds} \mu_0(d\thetab)}{\int_D e^{-\alpha \int_0^t V(\Thetab(s,\thetab),[\mu_s]) ds}}\right)^2\\
    &\quad + \frac{\int_D \dot \Thetab(t,\thetab) \cdot \nabla  V(\Thetab(t,\thetab),[\mu_t]) e^{-\alpha \int_0^t V(\Thetab(s,\thetab),[\mu_s]) ds} \mu_0(d\thetab)}{\int_D e^{-\alpha \int_0^t V(\Thetab(s,\thetab),[\mu_s]) ds} \mu_0(d\thetab)} \\
    &\quad + \frac{\int_{D\times D}  K(\Thetab(t,\thetab),\thetab') \partial_t \mu_t(d\thetab')  e^{-\alpha \int_0^t V(\Thetab(s,\thetab),[\mu_s]) ds} \mu_0(d\thetab)}{\int_D e^{-\alpha \int_0^t V(\Thetab(s,\thetab),[\mu_s]) ds} \mu_0(d\thetab)} 
    \end{aligned}
\end{equation}
Using \eqref{eq:16} to replace $\dot \Thetab(t,\thetab) $ by $- \nabla V(\Thetab(t,\thetab),[\mu_t])$ and \eqref{eq:15} to express these integral as expectations against~$\mu_t$ gives
\begin{equation}
    \begin{aligned}
    \frac{d}{dt} \bar V[\mu_t]  &= -\alpha \int_D |V(\thetab,[\mu_t])|^2  \mu_t(d\thetab)+\alpha \left( \int_D V(\thetab,[\mu_t])  \mu_t(d\thetab)\right )^2\\
        &\quad  - \int_D |\nabla  V(\thetab,[\mu_t])|^2  \mu_t(d\thetab)  
        -\int_{D\times D}  K(\thetab,\thetab') \partial_t \mu_t(d\thetab')   \mu_t(d\thetab)\\
        &= -\alpha \int_D \left(V(\thetab,[\mu_t])-\bar V[\mu_t]\right)^2  \mu_t(d\thetab) -  \int_D |\nabla  V(\thetab,[\mu_t])|^2  \mu_t(d\thetab)  \\
        &\quad  
        -\tfrac12 \frac{d}{dt} \int_{D\times D}  K(\thetab,\thetab') \mu_t(d\thetab')   \mu_t(d\thetab)
    \end{aligned}
\end{equation}
Therefore the terms at right hand side of \eqref{eq:Edecay} must be well-defined and we must also have
\begin{equation}
    \int_D |V(\thetab,[\mu_t])|^2\mu_t(d\thetab) < \infty, \qquad \int_D |\nabla V(\thetab,[\mu_t])|^2\mu_t(d\thetab)<\infty \qquad 
    \int_{D\times D}  K(\thetab,\thetab') \mu_t(d\thetab')   \mu_t(d\thetab)< \infty
\end{equation}
Since
$\mu_t \rightharpoonup \mu_* \in \mathcal{M}(D)$ by assumption, we can take the limit as $t\to\infty$ to deduce that
\begin{equation}
  \label{eq:17}
  \begin{aligned}
    \lim_{t\to\infty} \int_D V(\thetab,[\mu_t])\mu_t(d\thetab) & = \int_D
    V(\thetab,[\mu_*]) \mu_*(d\thetab) \\
    \lim_{t\to\infty} \int_D |V(\thetab,[\mu_t])|^2\mu_t(d\thetab) & = \int_D
    |V(\thetab,[\mu_*])|^2 \mu_*(d\thetab)\\
    \lim_{t\to\infty} \int_D |\nabla V(\thetab,[\mu_t])|^2\mu_t(d\thetab) & = \int_D
    |\nabla V(\thetab,[\mu_*])|^2 \mu_*(d\thetab)
  \end{aligned}
\end{equation}
We will use these properties below, along with
\begin{equation}
  \label{eq:12}
  V(\thetab,[\mu_t]) \to V(\thetab,[\mu_*]) \quad \text{and} \quad
  \int_D K(\thetab,\thetab') \mu_t(d\thetab') \to \int_D
  K(\thetab,\thetab') \mu_*(d\thetab') \qquad \text{pointwise in $D$}
\end{equation}
which is require in order that both $\bar V[\mu_t]$ and $\mathcal{E}[\mu_t]$ be well-defined at all $t>0$ and in the limit as $t\to\infty$.

\medskip

With these preliminaries, we now recall that the argument given after
Theorem~\ref{th:global} implies that any fixed point $\mu_*$ of the
PDE~\eqref{eq:pde_control} must satisfy the first equation
in~\eqref{eq:euler_lagrange}. That is, we must have
\begin{equation}
  \label{eq:3pp}
  V(\thetab,[\mu_*]) = \bar V[\mu_*] \qquad \forall \thetab \in \supp \mu_*
\end{equation}
Therefore, to prove Theorem~\ref{th:global}, it remains to show that the second
equation in~\eqref{eq:euler_lagrange} must be satisfied as well. We
will argue by contradiction: Let $D_* = \supp \mu_*$, assume
$D_*^c\not=\emptyset$, and suppose that there exists a region
$N\subseteq D_*^c$ where $V(\thetab,[\mu_*]) < \bar V[\mu_*]$. If it
exists, this region must have nonzero Hausdorff measure in $D$ since,
by Assumption~\ref{as:interacting-case},
$V(\thetab,[\mu_t]) \in C^2(D)$ for all $t\ge0$ and
$V(\thetab,[\mu_*]) \in C^2(D)$. $V(\thetab,[\mu_*]) -\bar V[\mu_*]$ must also reach
a minimum value inside $D$ even if $D$ is open, for otherwise~(\ref{eq:16}) would
eventually carry mass towards infinity, which contradicts $\mu_t
\rightharpoonup \mu_*$. This implies that, if we pick 
$\delta\in(0, \bar V[\mu_*] -\min_{\thetab} V(\thetab,[\mu_*]))$ and let
\begin{equation}
  \label{eq:4pp}
  N_{\delta} = \{\thetab : \delta \le \bar V[\mu_*] -
  V(\thetab,[\mu_*])\} \subset N,
\end{equation}
then $N_\delta$ is not empty. Since $V(\thetab,[\mu_*])$ is twice
differentiable in $\thetab$, for $\delta$ close enough to
$\bar V[\mu_*] - \min_{\thetab}V(\thetab,[\mu_*])$, $N_\delta$ is also
compact and such that
\begin{equation}
    \label{eq:boundnablaV}
    \forall \thetab \in \partial N_\delta \ : \qquad |\nabla V(\thetab,[\mu_*])|>0.
\end{equation}
Given 
any solution $\mu_t$ of the PDE~\eqref{eq:pde_control} that is supposed to converge to $\mu_*$ as $t\to\infty$, consider
\begin{equation}
    f_\delta (t) =  \mu_t(N_\delta)
\end{equation}
Since $\mu_t$ is positive everywhere at any finite time, we must have
$f_\delta (t)>0$ for $t\in(0,\infty)$ However, since $\mu_t\to\mu_*$,
we must also have
\begin{equation}
  \label{eq:5pp}
  \lim_{t\to\infty} f_\delta (t) = 0.
\end{equation} 
From~\eqref{eq:pde_control}, $f_\delta(t)$ satisfies
\begin{equation}
  \label{eq:6pp0}
  \begin{aligned}
    \dot f_\delta(t) & =   \int_{\partial N_\delta} 
    \hat n \cdot \nabla
      V d\sigma_t - \alpha \int_{N_\delta}( V-\bar V) d\mu_t
  \end{aligned}
\end{equation}
where $\hat n(\thetab)$ is the inward pointing unit normal to
$\partial N_\delta$ at $\thetab$ and $\sigma_t$ is the probability
measure on $\partial N_\delta$ obtained by restricting $\mu_t$ on this
boundary: If $\phi_{\epsilon}\in C^\infty_c(D)$ is a sequence of test
functions with $\supp \phi_\epsilon = N_\delta$ and converging towards
the indicator set of $N_\delta$ as $\epsilon\to0$, $\sigma_t$ is
defined as
\begin{equation}
  \lim_{\epsilon\to0} \int_{N_\delta}\nabla \phi_\epsilon(\thetab) \cdot \nabla V(\thetab,[\mu_t]) \mu_t(d\thetab) = \int_{\partial N_\delta}\hat n(\thetab) \cdot \nabla V(\thetab,[\mu_t]) d\sigma_t(\thetab)
\end{equation} 
Since 
\begin{equation}
    \lim_{t\to\infty} \hat n(\thetab) \cdot \nabla
V(\thetab,[\mu_t]) = |\nabla V(\thetab,[\mu_*])| >0,
\end{equation} 
there exists $t_+>0$ such that 
\begin{equation}
    \forall t> t_+ \ : \qquad \int_{\delta N_\delta} 
    \hat n \cdot \nabla
      V d\nu_t >0.
\end{equation} 
Restricting ourselves to $t>t_+$, we therefore have
\begin{equation}
  \label{eq:6pp00}
  \begin{aligned}
    \dot f_\delta(t) >  - \alpha \int_{N_\delta}( V-\bar V) d\mu_t
  \end{aligned}
\end{equation}
Let us analyze the remaining integral in this equation.
Denoting $\tilde
V(\thetab,[\mu_t]) =  V(\thetab,[\mu_t]) - \bar
      V[\mu_t]$, we have
\begin{equation}
  \label{eq:7pp}
\begin{aligned}
  -\alpha \int_{N_\delta}  \tilde V(\thetab,[\mu_t]) \mu_t(d\thetab) &
  = -\alpha \int_{N_\delta} \tilde V(\thetab,[\mu_*])
  \mu_t(d\thetab)\\
  & \quad -\alpha \int_{N_\delta} 
   \left(\tilde V(\thetab,[\mu_t]) - 
   \tilde V(\thetab,[\mu_*]) \right)
  \mu_t(d\thetab)\\
  & \ge \alpha \delta f_\delta(t) -\alpha \int_{N_\delta}
   \left(\tilde V(\thetab,[\mu_t]) - 
   \tilde V(\thetab,[\mu_*]) \right)
  \mu_t(d\thetab)
\end{aligned}
\end{equation}
where we used the definition of $N_\delta$. Looking at the  last term,
we can assess its magnitude using
\begin{equation}
  \label{eq:8pp}
\begin{aligned}
  & \left|\int_{N_\delta} 
   \left(\tilde V(\thetab,[\mu_t]) - 
   \tilde V(\thetab,[\mu_*]) \right)
  \mu_t(d\thetab)\right|\\
  & \le \tfrac12 \int_{N_\delta}
  \left|\int_{D} K(\thetab,\thetab') \left(\mu_t(d\thetab')-\mu_*(d\thetab')\right)\right|
  \mu_t(d\thetab) 
  + \left|\bar V[\mu_t]-\bar V(\mu_*)\right| f_\delta(t)\\
  &  \le M(t) f_\delta(t)
\end{aligned}
\end{equation}
where (using the compactness of $N_\delta$)
\begin{equation}
  \label{eq:9pp}
  \begin{aligned}
    M(t) & = \max_{N_{\delta}} \left|\int_D K(\thetab,\thetab')
      (\mu_t(d\thetab')-\mu_*(d\thetab'))\right| + |\bar V[\mu_t]-\bar
      V(\mu_*)|< \infty
  \end{aligned}
\end{equation}
Summarizing, we have deduced that
\begin{equation}
  \label{eq:10pp}
  \dot f_\delta(t) > \alpha \delta f_\delta(t) +
  R(t)
\end{equation}
with
\begin{equation}
  \label{eq:11p}
  |R(t)| \le M (t)f_\delta(t)
\end{equation}
Since we work under the assumption that $\mu_t \rightharpoonup\mu_*$,
$M(t)$ must tend to $0$ as $t\to\infty$. As a result,
$\exists t_\delta>0$ such $\forall t>t_\delta$ we have $N(t)< \delta$,
which, from~\eqref{eq:10pp}, implies that
$\forall t > \max(t_+,t_\delta)$ we have $\dot f_\delta(t) >0$, a
contradiction with~\eqref{eq:5pp}. Therefore the only fixed points
accessible by the PDE~\eqref{eq:pde_control} are those for which both
equations in~\eqref{eq:euler_lagrange} hold, which proves the theorem.

\section{Proof of Theorem~\ref{th:local-int-rate}}
\label{sec:convergence-interacting}

Let $\mu_*=\lim_{t\to\infty} \mu_t $ be the stationary point reached
by the solution of~\eqref{eq:pde_control} and denote
$E(t)=\mathcal{E}[\mu_t]- \mathcal{E}[\mu_*]\ge 0$. Then
\begin{equation}
  \label{eq:29}
  \begin{aligned}
    \frac{d}{dt} E^{-1} & = - E^{-2}\int_D V
    \partial_t \mu_t \\
    & =  E^{-2} \int_D \left(|\nabla V|^2 +\alpha |V - \bar V|^2 \right) 
    d\mu_t\\
    & \ge  \alpha E^{-2} \int_D  |V - \bar V|^2 
    d\mu_t
  \end{aligned}
\end{equation}
where we used
$\int_D V^2 d\mu_t - \bar V^2 = \int_D|V-\bar V|^2d\mu_t$. By
convexity
\begin{equation}
  \label{eq:30}
  \begin{aligned}
    \mathcal{E}[\mu_*] & \ge \mathcal{E}[\mu] - \int_D V
    (d\mu-d\mu_*) \\
    & = \mathcal{E}[\mu] -\bar V+\int_D V d\mu_*\\
    & = \mathcal{E}[\mu] +\int_D (V-\bar V) d\mu_*
  \end{aligned}
\end{equation}
As a result
\begin{equation}
  \label{eq:6}
  0\le E \le \int_D (\bar V-V) d\mu_*
\end{equation}
and hence
\begin{equation}
  \label{eq:31}
  0 \le E^2 \le \left|\int_D (V-\bar V) d\mu_*\right|^2
\end{equation}
Using this inequality in~\eqref{eq:29} gives
\begin{equation}
  \label{eq:33}
  \frac{d}{dt} E^{-1} \ge \alpha \frac{\int_D|V-\bar V|^2d\mu_t}
  {\left|\int_D (V-\bar V) d\mu_*\right|^2}
\end{equation}
In Lemma~\ref{lem:boundbelow} below we show that $\exists t_+ >0$ such that
\begin{equation}
  \label{eq:boundratio1}
   \forall t> t_+ \ : \ \frac{\int_D|V-\bar V|^2d\mu_t}
  {\left|\int_D (V-\bar V) d\mu_*\right|^2} \ge C >0 
\end{equation}
As a result, $dE^{-1}/dt \ge \alpha$ for $t>t_+$.
Integrating this relation in time on $[t_0,t]$ with $t_+<t_0\le t$ gives
\begin{equation}
  \label{eq:34}
  E^{-1}(t) \ge E^{-1}(t) - E^{-1}(t_0) \ge \alpha C (t-t_0)
\end{equation}
and hence
\begin{equation}
  \label{eq:37}
  \lim_{t\to\infty} t E(t) \le (\alpha C)^{-1} 
\end{equation}
which proves the theorem.\hfill $\square$

Note that the proof only takes into account the effects of birth-death terms;
adding transport may accelerate the rate.

\begin{lemma}
  \label{lem:boundbelow} There exist $t_+>0$ such
  that~(\ref{eq:boundratio1}) holds.
\end{lemma}

\noindent
\textit{Proof:} Let $\nu_t = \mu_t - \mu_*$ and for future reference
note that $\nu_t$ is a signed measure on $D_*=\supp \mu_*$ but
$\nu_t \ge 0 $ on $D^*_c$. Denote
\begin{equation}
  \label{eq:1}
  V = V(\thetab,[\mu_t]), \qquad \bar V = \int_D
  V(\thetab,[\mu_t]) d\mu_t, \qquad V_* = V(\thetab,[\mu_*]), \qquad \bar V_* = \int_D
  V(\thetab,[\mu_*]) d\mu_* 
\end{equation}
We have
\begin{equation}
  \label{eq:5bbb}
  \begin{aligned}
    V & = F(\thetab) 
  + \int_{D} K(\thetab,\thetab')
    (\mu_*(d\thetab') + \nu_t (d\thetab)') \\
    & = V_* + \int_{D} K(\thetab,\thetab') \nu_t (d\thetab') 
  \end{aligned}
\end{equation}
and hence
\begin{equation}
  \label{eq:5bb}
  \begin{aligned}
    \int_D V d\mu_* =\bar V_* + \int_{D\times
      D} K(\thetab,\thetab') \nu_t (d\thetab')\mu_*(d\thetab) 
  \end{aligned}
\end{equation}
Recall that $V_* = \bar V_*$ on $\supp \mu_*$. As  a result
\begin{equation}
  \label{eq:5}
  \begin{aligned}
    \bar V & = \int_D F(\thetab) (\mu_*(d\thetab) + \nu_t
    (d\thetab)) + \int_{D\times D} K(\thetab,\thetab')
    (\mu_*(d\thetab) + \nu_t (d\thetab)) (\mu_*(d\thetab') + \nu_t
    (d\thetab'))\\
    & = \bar V_* + \int_D F(\thetab) \nu_t (d\thetab) + 2 \int_{D\times
      D} K(\thetab,\thetab') \mu_*(d\thetab) \nu_t (d\thetab') + \int_{D\times
      D} K(\thetab,\thetab') \nu_t(d\thetab) \nu_t (d\thetab')
  \end{aligned}
\end{equation}
We can combine these two equations to obtain
\begin{equation}
  \label{eq:4}
  \begin{aligned}
    \int_D (\bar V - V) d\mu_* &= \int_D F(\thetab) \nu_t (d\thetab)
    +\int_{D\times D} K(\thetab,\thetab') \mu_*(d\thetab) \nu_t
    (d\thetab') + \int_{D\times D} K(\thetab,\thetab') \nu_t(d\thetab)
    \nu_t (d\thetab')\\
    & = \int_D V_*d\nu_t +\int_{D\times D}
    K(\thetab,\thetab') \nu_t(d\thetab) \nu_t (d\thetab') \\
    & = \int_D (V_\star - \bar V_*) d\nu_t+\int_{D\times D}
    K(\thetab,\thetab') \nu_t(d\thetab) \nu_t (d\thetab') \\
    & = \int_{D_*^c} (V_\star - \bar V_*) d\nu_t+\int_{D\times D}
    K(\thetab,\thetab') \nu_t(d\thetab) \nu_t (d\thetab')
  \end{aligned}
\end{equation}
where we used
$\int_D \bar V_* d\nu_t = \bar V_* \int_D(d\mu_t -d\mu_*) =0$ to get
the penultimate equality and $V_\star - \bar V_* =0$ on $D_*$ to get
the last.

Proceeding similarly using again $V_* = \bar V_*$ on $\supp \mu_*$ as
well as $\int_D d\nu_t =\int_D (d\mu_t-d\mu_*) = 0$, we can also
obtain
\begin{equation}
  \label{eq:7a}
  \begin{aligned}
    \int_D|V-\bar V|^2 d\mu_*& = 
    \int_D \left(\int_DK(\thetab,\thetab')
      \nu_t(d\thetab')\right)^2\mu_*(d\theta)
    +R^2-2R \int_{D\times D}
    K(\thetab,\thetab') \nu_t(d\thetab')
    \mu_*(d\thetab)
  \end{aligned}
\end{equation}
and
\begin{equation}
  \label{eq:7}
  \begin{aligned}
    \int_D|V-\bar V|^2 d\mu_t& = \int_{D_*^c}|V_*-\bar V_*|^2 d\nu_t +
    \int_D \left(\int_DK(\thetab,\thetab') \nu_t(d\thetab')\right)^2
    (\mu_*(d\theta) +\nu_t(d\thetab))+R^2\\
    & -2R \int_{D^c_*}(V_*-\bar V_*) d\nu_t-2R \int_{D\times D}
    K(\thetab,\thetab') \nu_t(d\thetab')
    (\mu_*(d\thetab)+\nu_t(d\thetab))\\
    & +2 \int_{D_*^c\times D}
    (V_*-\bar V_*) \nu_t(d\thetab) K(\thetab,\thetab')
    \nu_t(d\thetab') 
  \end{aligned}
\end{equation}
where we denote
\begin{equation}
  \label{eq:8}
  \begin{aligned}
    R &= \bar V- \bar V_* \\
    & = \int_D F(\thetab) \nu_t (d\thetab) + 2 \int_{D\times D}
    K(\thetab,\thetab') \mu_*(d\thetab) \nu_t (d\thetab') +
    \int_{D\times D} K(\thetab,\thetab') \nu_t(d\thetab) \nu_t
    (d\thetab')
  \end{aligned}
\end{equation}

Let us now compare the square of~(\ref{eq:4}) to~(\ref{eq:7}). Since
$V_*-\bar V_* \ge0$ and $\nu_t \ge0$ on $D_*^c$, we have
\begin{equation}
  \label{eq:3}
  \int_{D^c_*}
  (V_\star-\bar V_*) d \nu_t \ge 0.
\end{equation}
We
distinguish two cases:

\medskip

\textbf{Case 1:} $\int_{D^c_*} (V_*-\bar V_*) d \nu_t > 0$ (which
requires $D_*^c\not=\emptyset$). Since $\nu_t \rightharpoonup 0$ as
$t\to\infty$ the last term in~(\ref{eq:5bb}) is higher order.  As a
result, for any $\delta >0$, $\exists t_1>0$ such that
\begin{equation}
  \label{eq:10}
  \forall t> t_1 \ :  \ \int_D (\bar V - V) d\mu_*\le (1+\delta) \int_{D^c_*}
  (V_\star-\bar V_*) d \nu_t 
\end{equation}
which also implies that (using again $\nu_t\ge 0$ on $D^c_*$)
\begin{equation}
  \label{eq:10t1}
  \begin{aligned}
    \forall t> t_1 \ : \left|\int_D (V - \bar V) d\mu_*\right|^2& \le
    (1+\delta)^2 \left|\int_{D^*} (V_*-\bar V_*) d
      \nu_t\right|^2\\
    & \le (1+\delta)^2 \nu_t(D^c_*) \int_{D^*} |V_\star-\bar V_*|^2 d
    \nu_t
  \end{aligned}
\end{equation}
Similarly, the first term at the right hand side of~(\ref{eq:7})
dominates all the other ones as $t\to\infty$ in the sense that, for
any $\delta >0$, $\exists t_2>0$ such that
\begin{equation}
  \label{eq:10t2}
  \forall t> t_2 \ :  \ \int_D|V-\bar V|^2 d\mu_t \ge (1-\delta)
  \int_{D_*^c}|V_*-\bar V_*|^2  d\nu_t 
\end{equation}
Taken together, (\ref{eq:10t1}) and (\ref{eq:10t2}) imply the statement
of the lemma with any $C>0$ (since $\nu_t(D^c_*)\to0$ as
$t\to\infty$). As a result $\lim_{t\to\infty} t E(t) =0$ in this case
since
$\int_D|V - \bar V|^2 d\mu_t/|\int_D (V - \bar V) d\mu_*|^2 \to
\infty$.

\medskip

\textbf{Case 2:} $\int_{D^c_*} (V_*-\bar V_*) d \nu_t = 0$
(i.e. $D_*^c=\emptyset$ or $V_* = \bar V_*$ on $D_*^c$ as well as
$D_*$). In this case it is easier to use \eqref{eq:7a} via the
inequality
\begin{equation}
  \label{eq:11}
  \left|\int_D(V-\bar V) d\mu_*\right|^2\le \int_D|V-\bar V|^2 d\mu_*
\end{equation}
We also have that
\eqref{eq:7} reduces to 
\begin{equation}
  \label{eq:7cc}
  \begin{aligned}
    \int_D|V-\bar V|^2 d\mu_t& = 
    \int_D \left(\int_DK(\thetab,\thetab') \nu_t(d\thetab')\right)^2
    (\mu_*(d\theta) +\nu_t(d\thetab))+R^2\\
    & -2R \int_{D\times D}
    K(\thetab,\thetab') \nu_t(d\thetab')
    (\mu_*(d\thetab)+\nu_t(d\thetab))\\
    & = 
    \int_D \left(\int_DK(\thetab,\thetab') \nu_t(d\thetab')\right)^2
    (\mu_*(d\thetab) +\nu_t(d\thetab))-R^2
  \end{aligned}
\end{equation}
where we use the fact that $R$ reduces to (using $V_* =\bar V_*$ and
$\int_DV_*d\nu_t = \bar V_* \int_D(d\mu_t-d\mu_*)=0$)
\begin{equation}
  \label{eq:8cc}
  \begin{aligned}
    R &= \int_D V_* d\nu_t + \int_{D\times D}
    K(\thetab,\thetab') \mu_*(d\thetab) \nu_t (d\thetab') +
    \int_{D\times D} K(\thetab,\thetab') \nu_t(d\thetab) \nu_t
    (d\thetab') \\
    &= 
    \int_{D\times D} K(\thetab,\thetab') \nu_t(d\thetab) (\mu_*(d\thetab')+\nu_t
    (d\thetab')) 
  \end{aligned}
\end{equation}
Since
$\int_{D} K(\thetab,\thetab') \nu_t(d\thetab')\not =0$ on $D_*$, the
leading order terms in $\int_D|V-\bar V|^2 d\mu_*$ and
$\int_D|V-\bar V|^2 d\mu_t$ are the same and given by
\begin{equation}
  \label{eq:7ccc}
  \begin{aligned}
     A = \int_D
    \left(\int_DK(\thetab,\thetab') \nu_t(d\thetab')\right)^2 d\mu_* -
    \left(\int_{D\times D} K(\thetab,\thetab') \nu_t(d\thetab')
      \mu_*(d\thetab)\right)^2 >0
  \end{aligned}
\end{equation}
That is, for any $\delta >0$, $\exists t_3>0$ such that
\begin{equation}
  \label{eq:10bb}
  \forall t> t_3 \ :  \ \int_D|V-\bar V|^2 d\mu_*\le (1+\delta) A,
  \qquad \int_D|V-\bar V|^2 d\mu_t\ge (1-\delta) A
\end{equation}
Together with~(\ref{eq:11}), this implies the statement of the lemma
with $C=1$.
\hfill $\square$

 \section{Proof of Propositions~\ref{th:lln} and \ref{th:clt}} 
\label{app:flucts}

Here we give formal proofs Propositions~\ref{th:lln} and \ref{th:clt}
using tools from the theory of measure-valued Markov
processes~\cite{dawson2006measure}.

To begin, recall that the evolution of
$\mu^{(n)}_t=n^{-1} \sum_{i=1}^n \delta_{\thetab_i(t)}$ is Markovian
since that of the particles $\thetab_i(t)$ is and these particles are
interchangeable. To study this measure-valued Markov process and in
particular analyze its properties when $n \to\infty$, it is useful to
write its infinitesimal generator, i.e. the operator whose action on
a functional $\Phi:\mathcal{M}(\RR^k) \to \RR$ evaluated on
$\mu^{(n)}$ is defined via
\begin{equation}
  \label{eq:generatordef}
  ({\mathcal L}_n\Phi)[\mu^{(n)}] = \lim_{t\to0+} t^{-1} \left( \EE^{\mu_0^{(n)}=\mu^{(n)}}
    \Phi[\mu_t^{(n)}] - \Phi[\mu^{(n)}]\right) 
\end{equation}
where $\EE^{\mu_0^{(n)}=\mu^{(n)}}$ denotes the expectation along the
trajectory $\mu^{(n)}_t$ taken conditional on $\mu_0^{(n)}=\mu^{(n)}$
for some given $\mu^{(n)}$. To compute the limit
in~(\ref{eq:generatordef}), notice that if particle $\thetab_i(t)$
gets killed at time $t$ and particle $\thetab_j(t)$ gets duplicated,
the changes this induces on $\mu^{(n)}_t$ is 
\begin{equation}
\mu_t^{(n)} = \mu^{(n)}_{t-} + n^{-1} \left(\delta_{\thetab_j} - \delta_{\thetab_i} \right).
\end{equation}
where $\mu^{(n)}_{t-}= \lim_{\epsilon \to 0+} \mu^{(n)}_{t-\epsilon}$
Similarly if particle $\thetab_i(t)$ gets duplicated at time $t$ and
particle $\thetab_j(t)$ gets killed, the change this induces on
$\mu^{(n)}_t$ is
\begin{equation}
  \mu_t^{(n)} = \mu^{(n)}_{t-} - n^{-1} \left(\delta_{\thetab_j} - \delta_{\thetab_i} \right).
\end{equation}
A particle swap occurs with rates dictated by $\tilde V$, so we define
\begin{equation}
  \label{eq:swap}
  \mu^{(n)}_t\{\thetab\leftrightarrow\thetab'\} = \mu^{(n)}_{t} + n^{-1} \sigma(\thetab) \left(\delta_{\thetab} - \delta_{\thetab'} \right)
\end{equation}
where $\sigma(\thetab_i) = \sign \tilde V(\thetab_i)$ determines the direction of the swap.
If we account for the rate at which these events occur, as well as the
effect of transport by GD, we can explicitly compute the generator
defined in~(\ref{eq:generatordef}) and arrive at the expression
\begin{equation}
  \label{eq:generator0inter}
  \begin{aligned}
    ({\mathcal L}_n\Phi)[\mu^{(n)}] &= -\frac1n \sum_{i=1}^n \int_D
    \nabla V(\thetab_i,[\mu^{(n)}])
    \delta_{\thetab_i}(d\thetab) \cdot \nabla_{\thetab_i} D_{\mu^{(n)}} \Phi(\thetab_i)\\
    &\quad + \frac{\alpha}{n} \sum_{i,j=1}^n \int_{D\times D} |\tilde
    V(\thetab_i)|\delta_{\thetab_i}(d\thetab)
    \delta_{\thetab_j}(d\thetab') \left( \Phi[\mu^{(n)}_t\{\thetab_i\leftrightarrow\thetab_j\}]
      - \Phi[\mu^{(n)}] \right)
 \end{aligned}
\end{equation}
where  the
functional derivative $D_\mu \Phi$
is the function from $D$ to
$\RR$ defined via: for any $\omega\in \mathcal{M}_s(D)$, the space
of signed distributions such that $\int_{D} \omega(d\thetab) =0$,
\begin{equation}
  \label{eq:fder}
  \lim_{\eps\to0} \eps^{-1} \left(  \Phi[\mu+\epsilon \omega
    ]-\Phi[\mu]\right)  =
  \int_{D} D_\mu \Phi (\thetab)\omega(d\thetab) 
\end{equation}
We can use the properties of the Dirac distribution to rewrite the
generator in~(\ref{eq:generator0inter}) as
 \begin{equation}
  \label{eq:generator1inter}
  \begin{aligned}
    ({\mathcal L}_n\Phi)[\mu^{(n)}] &= -\int_D \nabla V(\thetab,[\mu^{(n)}])
    \mu^{(n)}(d\thetab) \cdot  \nabla D_{\mu^{(n)}} \Phi(\thetab)\\
    & + n \alpha \int_{D\times D} | \tilde V(\thetab,
    [\mu^{(n)}])|\mu^{(n)}(d\thetab) \mu^{(n)}(d\thetab') \left(
      \Phi[\mu^{(n)}_t\{\thetab\leftrightarrow\thetab'\}] - \Phi[\mu^{(n)}] \right)
  \end{aligned}
\end{equation}
and $\sigma$ in~\eqref{eq:swap} is evaluated on
\begin{equation}
  \label{eq:Fcenteredmuinter}
  \tilde V(\thetab,[\mu]) = F(\thetab) + \int_D K(\thetab,\thetab')
  \mu(d\thetab') - \int_{D} \left( F(\thetab') + \int_D K(\thetab',\thetab'')
    \mu(d\thetab'') \right) \mu(d\thetab').
\end{equation}
The operator in~(\ref{eq:generator1inter}) is now defined for any
$\mu\in \mathcal{M}(D)$, and we will use it in this form in our
developments below.

The generator~(\ref{eq:generator1inter}) can be used to write an
evolution equation for the expectation of functionals evaluated on
$\mu^{(n)}_t$. That is, if we define
\begin{equation}
  \label{eq:observable}
  \Phi_t[\mu^{n}] = \EE^{\mu_0^{(n)}=\mu^{(n)}} \Phi[\mu_t^{n}]
\end{equation}
then this time-dependent functional satisfies the backward Kolmogorov
equation (BKE)
\begin{equation}
  \label{eq:BKE}
  \partial_t \Phi_t[\mu^{n}] = ({\mathcal
    L}_n\Phi_t)[\mu^{(n)}],\qquad  \Phi_{t=0}[\mu^{n}] =  \Phi[\mu^{n}].
\end{equation}
The proof of Proposition~\ref{th:lln} is based on analyzing the
properties of this equation in the limit as $n\to\infty$, which we expand upon in
Appendix~\ref{app:lln}. The proof of Proposition~\ref{th:clt} is
based on writing a similar equation for an extended process in which we magnify the dynamics of $\mu^{(n)}_t$ around its limit,
as shown in Appendix~\ref{app:clt}.

\subsection{Proof of Proposition~\ref{th:lln}}
\label{app:lln}

If  we take the limit of
$({\mathcal L}_n{\Phi})[\mu^{(n)}]$ as $n\to\infty$ on a sequence such
that $\mu^{(n)}\rightharpoonup \mu$, we deduce that
$({\mathcal L}_n{\Phi})[\mu^{(n)}] \to ({\mathcal L}{\Phi})[\mu]$ with
\begin{equation}
  \label{eq:limgeninter}
  \begin{aligned} 
    ({\mathcal L}{\Phi})[\mu] &= -\int_D \nabla V(\thetab,[\mu])
    \mu(d\thetab) \cdot \nabla_{\thetab} D_{\mu} \Phi(\thetab) -\alpha
    \int_{D} \tilde V(\thetab,[\mu]) \mu(d\thetab) D_\mu\Phi(\thetab)
  \end{aligned}
\end{equation}
Correspondingly, in this limit the BKE~(\ref{eq:BKE}) becomes
\begin{equation}
  \label{eq:BKElln}
  \partial_t \Phi_t[\mu] = ({\mathcal
    L}\Phi_t)[\mu],\qquad  \Phi_{t=0}[\mu] =  \Phi[\mu].
\end{equation}
Since~(\ref{eq:limgeninter}) is precisely the generator of process
defined by the PDE~\eqref{eq:pde_control}, this shows that, if
$\mu^{(n)}_{t=0} = \mu^{(n)}\rightharpoonup \mu$ as $n\to \infty$,
then
\begin{equation}
  \label{eq:limobs}
  \lim_{n\to\infty} \Phi_t[\mu^{(n)}]  = \Phi_t[\mu] \quad
  \Leftrightarrow \quad \lim_{n\to\infty} \EE^{\mu_0^{(n)}=\mu^{(n)}}
    \Phi[\mu_t^{n}] = \Phi[\mu_t]
\end{equation}
where $\mu_t$ solves the PDE~\eqref{eq:pde_control} for the initial
condition $\mu_{t=0} = \mu$. This proves the weak version of the LLN
stated in Proposition~\ref{th:lln}.

\subsection{Proof of Proposition~\ref{th:clt}}
\label{app:clt}

To quantify the fluctuations around the LLN, let $\mu_t$ be the limit
of $\mu^{(n)}_t$ (i.e. the solution to the PDE~\eqref{eq:pde_control})
and define
\begin{equation}
  \label{eq:discrepency}
    \omega_t^{(n)} = \sqrt{n} \left(\mu^{(n)}_t - \mu_t \right) \in \mathcal{M}_s(D)
\end{equation}
We can write down the generator of the joint process
$(\mu_t,\omega_t^{(n)})$. 
To do so, we consider its action on a functional,
$\hat \Phi: \mathcal{M}(D)\times \mathcal{M}_s(D)\to \RR$ is
given by (using $\mu^{(n)}= \mu + n^{-1/2} \omega^{(n)}$)
\begin{equation}
  \label{eq:generator2}
  \begin{aligned} 
    & (\Hat{\mathcal L}_n{\hat\Phi})[\mu,\omega^{(n)}]\\
    &= n^{1/2}\int_D \nabla V(\thetab, [\mu+n^{-1/2} \omega^{(n)}]))
    \left(\mu(d\thetab)+n^{-1/2} \omega^{(n)}(d\thetab) \right) \cdot
    \nabla
    D_{\omega^{(n)}}\hat \Phi(\thetab)\\
    & + n \alpha \int_{D\times D} \sigma(\thetab,[\mu+n^{-1/2}
    \omega^{(n)}]) \tilde V(\thetab, [\mu+n^{-1/2}
    \omega^{(n)}])\left(\mu(d\thetab) + n^{-1/2}
      \omega^{(n)}(d\thetab)\right) \left(\mu(d\thetab')
      + n^{-1/2}\omega^{(n)}(d\thetab')\right)\\
    & \qquad \qquad \times \left( \hat \Phi[\mu,\omega^{(n)} +
      n^{-1/2} \sigma(\thetab, [\mu+n^{-1/2} \omega^{(n)}])
      (\delta_{\thetab'}-\delta_{\thetab})]
      - \hat \Phi[\mu,\omega^{(n)}] \right) \\
    & -\int_D \nabla V(\thetab, [\mu])) \mu(d\thetab) \cdot
    \nabla_{\thetab} \left(D_\mu \hat \Phi(\thetab) - n^{1/2}
      D_{\omega^{(n)}}\hat \Phi(\thetab)\right) \\
    & - \alpha \int_{D} \tilde V(\thetab,[\mu]) \mu(d\thetab) \left(
      D_\mu \hat \Phi(\thetab) - n^{1/2} D_{\omega^{(n)}}\hat
      \Phi(\thetab)\right).
  \end{aligned}
\end{equation}

Proceeding similarly as we did to derive~(\ref{eq:limgeninter}), we
can take the limit of
$(\hat {\mathcal L}_n{\hat\Phi})[\mu,\omega^{(n)}]$ as $n\to\infty$ on
a sequence such that
$\omega^{(n)}\rightharpoonup \omega \in \mathcal{M}_s(D)$. A direct
calculation using $\int_{D}\tilde V(\thetab,[\mu]) d \mu=0$,
$\int_{D} d\omega = 0$, and
$\int_{D} \tilde V(\thetab,[\mu]) d \mu =1$ indicates that
$(\hat {\mathcal L}_n{\hat\Phi})[\mu,\omega^{(n)}]\to (\hat {\mathcal
  L}{\hat\Phi})[\mu,\omega]$ with
\begin{equation}
    \label{eq:limgen2}
    \begin{aligned} 
      (\hat {\mathcal L}{\hat\Phi})[\mu,\omega] &= -\int_D \nabla
      V(\thetab,[\mu]) \omega(d\thetab) \cdot \nabla
      D_{\omega} \Phi(\thetab) - \int_{D\times D} \nabla
      K(\thetab,\thetab') \omega(d\thetab') \mu(d\thetab) \cdot
      \nabla D_{\omega} \Phi(\thetab)\\
      & \quad -\alpha \int_{D} \tilde V(\thetab,[\mu])
      \omega(d\thetab) D_\omega\hat \Phi(\thetab) -\alpha
      \int_{D\times D} K(\thetab,\thetab') \omega(d\thetab')
      \mu(d\thetab)
      D_\omega \hat \Phi(\thetab) \\
      &\quad + \alpha \int_{D\times D} \tilde V(\thetab',[\mu])
      \omega(d\thetab') \mu(d\thetab)
      D_\omega\hat \Phi(\thetab)\\
      &\quad + \alpha \int_{D\times D\times D} K(\thetab',\thetab'')
      \omega(d\thetab') \mu(d\thetab'') \mu(d\thetab)
      D_\omega \hat \Phi(\thetab)\\
      &\quad+ \alpha \int_{D\times D} |\tilde V(\thetab,[\mu])|
      \mu(d\thetab) \mu(d\thetab') \left(D_\omega^2\hat
        \Phi(\thetab,\thetab) +D_\omega^2\hat \Phi(\thetab',\thetab')
        - 2 D_\omega^2\hat \Phi(\thetab,\thetab')\right)\\
      &\quad -\int_D \nabla V(\thetab,[\mu]) \mu(d\thetab) \cdot
      \nabla_{\thetab} D_{\mu} \Phi(\thetab) -\alpha \int_{D} \tilde
      V(\thetab,[\mu]) \mu(d\thetab) D_\mu\Phi(\thetab)
    \end{aligned}
\end{equation}
where the second order functional derivative $D^2_{\mu}\hat \Phi$ is
the function from $D\times D$ to $\RR$ defined via: for any
$\nu,\nu' \in \mathcal{M}_s(D)$,
\begin{equation}
  \label{eq:fctder2}
  \begin{aligned}
    & \lim_{\eps,\eps'\to0} (\eps\eps')^{-1} \left(\hat \Phi[\mu+\eps
      \nu +\eps '\nu',\omega] -\hat \Phi[\mu+\eps \nu,\omega]-\hat
      \Phi[\mu +\eps '\nu',\omega]+ \hat \Phi[\mu,\omega]\right)\\
    & \qquad = \int_{D\times D} D^2_\mu \hat \Phi(\thetab,\thetab')
    \nu(d\thetab)\nu(d\thetab'),
  \end{aligned}
\end{equation}
and similarly for $D^2_{\omega}\hat \Phi$.  
The operator in $\mu$
in~(\ref{eq:generator2}) is the same as in~(\ref{eq:limgeninter}),
confirming the LLN; the operator in~$\omega$ is a second order
operator, i.e. it is the generator of a stochastic differential
equation. That is, we have established that, as $n\to\infty$,
\begin{equation}
  \omega^{(n)}_t \equiv \sqrt{n} \left(\mu^{(n)}_t - \mu_t \right)
  \rightharpoonup \omega_t \qquad \text{in law}
\end{equation}
where $\omega_t(d\thetab)$ is Gaussian random distribution whose
equation can be obtained from the generator in~(\ref{eq:limgen2}) Formally
\begin{equation}
  \label{eq:omegainter}
  \begin{aligned}
    \partial_t \omega_t & = \nabla \cdot\left( \nabla
      V(\thetab,[\mu_t]) \omega_t + \int_{D}\nabla
      K(\thetab,\thetab') \omega_t(d\thetab') \mu_t\right)\\
    & \quad -\alpha \tilde V(\thetab,[\mu_t]) \omega_t
    - \alpha \int_D K(\thetab,\thetab') \omega_t(d\thetab')
    \mu_t \\
    & \quad + \alpha (\int_{D\times D}
    K(\thetab',\thetab'') \mu_t(d\thetab') \omega_t(d\thetab'')) \mu_t
    + \sqrt{2} \eta(t),
  \end{aligned}
\end{equation}
where $\eta(t)$ is a white-noise term with covariance consistent
with~\eqref{eq:limgen2}:
\begin{equation}
\begin{aligned}
  \EE \eta(t)\eta(t') &=
  \alpha | \tilde V(\thetab,[\mu_t]) | \mu_t(d\thetab)
  \delta_{\thetab} (d\thetab') \delta(t-t')\\
  &-\alpha \left(|\tilde V(\thetab,[\mu_t])|
    + |\tilde V(\thetab',[\mu_t])|\right) \mu_t(d\thetab) \mu_t(d\thetab') \delta(t-t')
    \end{aligned}
\end{equation}
Since $\omega_t$ is Gaussian with zero mean, all its information is
contained in its covariance
$\Sigma_t(d\thetab,d\thetab') = \EE \omega_t(d\thetab)
\omega_t(d\thetab')$, for which we can derive the equation
\begin{equation}
  \label{eq:covarianceinter}
  \begin{aligned}
    \partial_t \Sigma_t & = \nabla_{\thetab} \cdot\left( \nabla
      V(\thetab,[\mu_t]) \Sigma_t + \int_{D}\nabla
      K(\thetab,\thetab'') \Sigma_t(d\thetab,d\thetab'')
      \mu_t(d\thetab)\right) \\
    &\quad + \nabla_{\thetab'} \cdot\left( \nabla
      V(\thetab',[\mu_t]) \Sigma_t +\int_{D} \nabla
      K(\thetab',\thetab'') \Sigma_t(d\thetab',d\thetab'')
      \mu_t(d\thetab')\right) \\
    &\quad-\alpha \left(\tilde V(\thetab,[\mu_t])
      +\tilde V(\thetab',[\mu_t])\right) \Sigma_t \\
    & \quad - \alpha \mu_t(d\thetab) \int_{D} K(\thetab,\thetab'')
    \Sigma_t(d\thetab'',d\thetab') -\alpha \mu_t(d\thetab')
    \int_{D} K(\thetab',\thetab'') \Sigma_t(d\thetab'',d\thetab)\\
    & \quad + \alpha \mu_t(d\thetab) \int_{D} \tilde V(\thetab'',[\mu_t])
    \Sigma_t(d\thetab'',d\thetab) +\alpha \mu_t(d\thetab')
    \int_{D} \tilde V(\thetab'',[\mu])
    \Sigma_t(d\thetab'',d\thetab')\\
    & \quad + \alpha \mu_t(d\thetab) \int_{D\times D}
    K(\thetab''',\thetab'') \mu_t(d\thetab''')
    \Sigma_t(d\thetab'',d\thetab') +\alpha \mu_t(d\thetab')
    \int_{D} K(\thetab''',\thetab'') \mu_t(d\thetab''')\Sigma_t(d\thetab'',d\thetab)\\
    & \quad + \alpha |\tilde V(\thetab,[\mu_t])|\mu_t(d\thetab)
    \delta_{\thetab} (d\thetab') -\alpha (|\tilde
    V(\thetab,[\mu_t])|+ |\tilde V(\thetab',[\mu_t])|)
    \mu_t(d\thetab) \mu_t(d\thetab')
  \end{aligned}
\end{equation}
This equation should also be interpreted in the weak sense by testing
it against some $\phi\in C_c^\infty(D\times D)$, and it can be seen
that it conserves mass  in the sense that
$\Sigma_t(d\thetab,D) = \Sigma_t(D,d\thetab') =0$ for all
$t>0$ since this is true initially and
$\partial_t \Sigma_t(d\thetab,D)
=\partial_t\Sigma_t(D,d\thetab') =0$.

We can also analyze the effect of the fluctuations at long times.
Since $|\tilde V(\thetab,[\mu_t])|\mu_t(d\thetab)\rightharpoonup 0$ as
$t\to\infty$, the noise terms in~(\ref{eq:omegainter}) and
(\ref{eq:covarianceinter}) converge to zero---a property we refer to
as self-quenching---and these equations reduce respectively
to \begin{equation}
  \label{eq:omegainterlt}
  \begin{aligned}
    \partial_t \omega_t & = \nabla \cdot\left( \nabla
      V(\thetab,[\mu_*]) \omega_t + \int_{D} \nabla
      K(\thetab,\thetab') \omega_t(d\thetab') \mu_*\right)\\
    & \quad -\alpha \tilde V(\thetab,[\mu_*]) \omega_t
    - \alpha \int_D K(\thetab,\thetab') \omega_t(d\thetab')
    \mu_* \\
    & \quad + \alpha ( \int_D V(\thetab',[\mu_*])
    d\omega_t(\thetab')) \mu_* + \alpha (\int_{D\times D}
    K(\thetab',\thetab'') \mu_*(d\thetab') \omega_t(d\thetab'')) \mu_*
  \end{aligned}
\end{equation}
and
\begin{equation}
  \label{eq:covarianceinterlt}
  \begin{aligned}
    \partial_t \Sigma_t & = \nabla_{\thetab} \cdot\left( \nabla
      V(\thetab,[\mu_*]) \Sigma_t + \int_{D} \nabla
      K(\thetab,\thetab'') \Sigma_t(d\thetab,d\thetab'')
      \mu_*(d\thetab)\right) \\
    & \quad+ \nabla_{\thetab'} \cdot\left( \nabla
      V(\thetab',[\mu_*]) \Sigma_t + \int_{D} \nabla
      K(\thetab',\thetab'') \Sigma_t(d\thetab',d\thetab'')
      \mu_*(d\thetab')\right) \\
    &\quad-\alpha \left(\tilde V(\thetab,[\mu_*])
      +\tilde V(\thetab',[\mu_*])\right) \Sigma_t \\
    & \quad - \alpha \mu_t(d\thetab) \int_{D} K(\thetab,\thetab'')
    \Sigma_t(d\thetab'',d\thetab') -\alpha \mu_*(d\thetab')
    \int_{D} K(\thetab',\thetab'') \Sigma_t(d\thetab'',d\thetab)\\
    & \quad + \alpha \mu_*(d\thetab) \int_{D} \tilde V(\thetab'',[\mu])
    \Sigma_t(d\thetab'',d\thetab) +\alpha \mu_*(d\thetab')
    \int_{D} \tilde V(\thetab'',[\mu_*])
    \Sigma_t(d\thetab'',d\thetab')\\
    & \quad + \alpha \mu_*(d\thetab) \int_{D\times D^2}
    K(\thetab''',\thetab'') \mu_*(d\thetab''')
    \Sigma_t(d\thetab'',d\thetab') +\alpha \mu_*(d\thetab')
    \int_{D^2} K(\thetab''',\thetab'') \mu_*(d\thetab''')\Sigma_t(d\thetab'',d\thetab)
  \end{aligned}
\end{equation}
Since $\tilde V(\thetab,[\mu_*]) \ge 0$, the fixed points of these
equations are $\omega_t=0$ and $\Sigma_t=0$. That is, the effect of
the fluctuations disappear as $t\to\infty$, and in particular they do
not impede in the particle system the convergence observed at mean
field level.

\end{document}